\pdfoutput=1
\documentclass{article}

\usepackage[utf8]{inputenc} %
\usepackage[T1]{fontenc}    %
\usepackage{hyperref}       %
\usepackage{booktabs}       %
\usepackage{amsfonts}       %
\usepackage{amsmath}
\usepackage{amssymb}
\usepackage{pifont}
\usepackage[dvipsnames]{xcolor}

\newcommand{\cmark}{\textcolor{OliveGreen}{\ding{51}}}
\newcommand{\xmark}{\textcolor{BrickRed}{\ding{55}}}

\usepackage{dsfont}
\usepackage{nicefrac}       %
\usepackage{microtype}      %
\usepackage{siunitx}
\usepackage{color}
\usepackage{graphicx}
\usepackage{tabularx}
\usepackage{caption}
\usepackage[english]{babel}
\usepackage{placeins}
\usepackage{subfig}
\usepackage{algorithm}
\usepackage{algorithmic}

\usepackage[letterpaper, total={6in, 9in}]{geometry}

\usepackage[square,sort,comma,numbers]{natbib}
\usepackage{authblk}

\graphicspath{{./figs/}}

\title{\textbf{DPSOM: Deep Probabilistic Clustering \\ with Self-Organizing Maps}}

\author[1]{Laura Manduchi}
\author[1]{Matthias H\"user}
\author[1]{Julia Vogt}
\author[1]{\\Gunnar R\"atsch}
\author[1]{Vincent Fortuin} 

\affil[1]{Department of Computer Science, ETH Z\"urich, Universit\"atsstrasse 6, 8092 Z\"urich, Switzerland}

\date{}

\begin{document}

\maketitle

\begin{abstract}

Generating interpretable visualizations from complex data is a common problem in many applications. Two key ingredients for tackling this issue are clustering and representation learning. However, current methods do not yet successfully combine the strengths of these two approaches. Existing representation learning models which rely on latent topological structure such as self-organising maps, exhibit markedly lower clustering performance compared to recent deep clustering methods. To close this performance gap, we (a) present a novel way to fit self-organizing maps with probabilistic cluster assignments (PSOM), (b) propose a new deep architecture for probabilistic clustering (DPSOM) using a VAE, and (c) extend our architecture for time-series clustering (T-DPSOM), which also allows forecasting in the latent space using LSTMs. We show that DPSOM achieves superior clustering performance compared to current deep clustering methods on MNIST/Fashion-MNIST, while maintaining the favourable visualization properties of SOMs. On medical time series, we show that T-DPSOM outperforms baseline methods in time series clustering and time series forecasting, while providing interpretable visualizations of patient state trajectories and uncertainty estimation.

\end{abstract} %
\section{Introduction}

Clustering is one of the most natural ways for retrieving interpretable information from raw data. Long-established methods such as k-means \citep{macqueen1967} and Gaussian 
Mixture Models \citep{Bishop:2006:PRM:1162264} represent the cornerstone of cluster analysis. Their applicability, however, is
constrained to simple data and their performance is limited in high-dimensional, complex, real-world data sets, which do not exhibit a clustering-friendly structure.
To overcome this issue, dimensionality reduction, such as PCA \citep{Jolliffe2011}, has been successfully applied to obtain a low-dimensional representation which is more suited for clustering. Recently, deep neural networks (DNNs) such as Autoencoders (AEs), %
Variational Autoencoders (VAEs) and Generative Adversarial Networks (GANs) \citep{2013arXiv1312.6114K, 2014arXiv1406.2661G}, have been used in combination with clustering methods to substantially increase their clustering performance \citep{DBLP:journals/corr/XieGF15}.
Indeed, the compressed latent representation generated by these models has been proven to ease the clustering process \citep{DBLP:journals/corr/abs-1801-07648}.
Although very successful, most of these methods do not investigate the relationship
between clusters.
Moreover, the clustered feature points lie in a high-dimensional latent space that cannot be easily visualized or interpreted by humans. 

In contrast, the Self-Organizing Map (SOM) \citep{58325} is a clustering method that provides such an interpretable representation. 
It produces a low-dimensional (typically 2-dimensional), discretized representation of the input space by inducing a flexible neighbourhood structure over the clusters.
Alas, its performance strongly depends on the complexity of the data sets used and, similarly to other classical clustering methods, it usually performs poorly on complex high-dimensional data. 
While the SOM is particularly effective for data visualization \citep{7280357}, only few methods have attempted to combine it with DNNs.  %
Moreover, as we will show in Section \ref{sec:experiments}, their performances are lower compared to modern clustering methods.

To address the above issues, we propose a novel way of fitting SOMs with probabilistic cluster assignments, which we call Probabilistic SOM (PSOM).
We moreover extend this PSOM to a deep architecture, the Deep Probabilistic SOM (DPSOM), which jointly trains a VAE and a PSOM to achieve an interpretable discrete representation while exhibiting state-of-the-art clustering performance. 
Instead of hard assignments of data points to clusters, our model uses centroid-based probability distributions. It minimizes their 
Kullback-Leibler divergence against auxiliary target distributions, while enforcing a SOM-friendly space.
To highlight the importance of an interpretable representation for temporal applications, we further extend this model to support time series,
yielding the temporal DPSOM (T-DPSOM). We discuss related work in Section~\ref{sec:related_work} and describe our model in Section~\ref{sec:model}. Extensive evidence of the superior clustering performance of both models, on MNIST/Fashion-MNIST images as well as medical time series, is presented in Section \ref{sec:experiments}.

Our main contributions are:

\begin{itemize}
	\item PSOM, a novel way of fitting SOMs using probabilistic cluster assignments.
	\item DPSOM, an architecture for deep clustering, yielding an interpretable discrete representation through the combination of a PSOM with a VAE.
	\item T-DPSOM, an extension of this architecture to time series, improving clustering performance on this data type and enabling temporal forecasting.
	\item A thorough empirical assessment of our proposed models, showing superior performance on benchmark tasks and medical time
	      series from the intensive care unit.
\end{itemize} %
\section{Related Work}
\label{sec:related_work}

\begin{table*}[t]
\caption{Overview of related approaches and our proposed methods.}
\label{table:overview_methods}
\vspace{-0.2cm}
\begin{center}
\resizebox{\linewidth}{!}{
\begin{tabular}{lcccc}\toprule
Model & SOM structure & Probabilistic & Clustering performance & Temporal model
\\ \midrule
DEC / IDEC \citep{DBLP:journals/corr/XieGF15, ijcai2017-243} & \xmark & \cmark & \cmark & \xmark \\
SOM-VAE \citep{DBLP:journals/corr/abs-1806-02199} & \cmark & \xmark & \xmark & \cmark \\
DESOM \citep{Forest2019} & \cmark & \xmark & \xmark & \xmark \\
DPSOM (ours) & \cmark & \cmark & \cmark & \xmark \\
T-DPSOM (ours) & \cmark & \cmark & \cmark & \cmark \\
\bottomrule
\end{tabular}
}
\end{center}
\vspace{-0.3cm}
\end{table*}

\paragraph{Self-organizing maps.}
Self-organizing maps have been widely used as a means to visualize information from large amounts of 
data \citep{7008682}. They can be seen as a form of clustering in which the centroids are connected by a topological neighborhood 
structure \citep{10.1007/978-3-540-48247-5_9}. Since their inception \citep{58325}, several variants have been 
proposed to enhance their performance and scope. The adaptive subspace SOM, ASSOM \citep{Kohonen1995TheAS}, for example, combines 
PCA and SOMs to map data into a reduced feature space.  \citet{TOKUNAGA200982} combine SOMs with 
multi-layer perceptrons to obtain a modular network. \citet{7280357} proposed the Deep SOM (DSOM), an architecture composed of 
multiple layers similar to deep neural networks.  
Although there exist several methods tailored to representation learning on time series \citep{franceschi2019unsupervised, fortuin2019deep, fortuin2019multivariate}, only few models present extensions of the SOM optimized for temporal data. Examples are the Temporal Kohonen map 
\citep{Chappell:1993:TKM:154879.154890}, its improved version Recurrent SOM \citep{10.1007/978-3-540-45240-9_1}, as well as 
Recursive SOM \citep{VOEGTLIN2002979}. Probabilistic versions of SOM include \citep{lopez2010probabilistic, cheng2009model}
as well as the generative topographic map \citep{bishop1998gtm}.

\paragraph{Deep clustering.}
Recent works on clustering analysis have shown that using deep neural networks (DNNs) in combination with clustering algorithms greatly increases the clustering performance  \citep{IEEE:journal2018Min,DBLP:journals/corr/abs-1801-07648}.
DNNs are used in that case to embed the data set into a space which is more suited for clustering.
\citet{DBLP:journals/corr/XieGF15} proposed DEC, a method that sequentially applies embedding learning using Stacked Autoencoders (SAE) and the Cluster Assignment Hardening method on the obtained representations. An improvement of this architecture, IDEC
\citep{ijcai2017-243}, includes the decoder network of the SAE in the learning process, so that training is affected by 
both the clustering loss and the reconstruction loss. Similarly, DCN \citep{DBLP:journals/corr/YangFSH16} combines a k-means 
clustering loss with the reconstruction loss of SAE to obtain an end-to-end architecture that jointly trains representations and
clustering. These models achieve state-of-the-art clustering performance, but they do not investigate the relationship among clusters.
An exception is the work by \citet{DBLP:journals/corr/abs-1803-05206}, which presents an unsupervised method that learns latent embeddings and discovers a multi-faceted clustering structure. However, these relationships between clusters do not provide a latent space that can be easily interpreted and which would ease the process of visual reasoning.

\paragraph{Deep SOM-based models.}
While there exist previous efforts to endow VAEs with a hierarchical latent space \citep{vikram2018loracs, goyal2017nonparametric},
to the best of our knowledge, only two models used deep generative models in combination with a SOM structure in the latent space. %
The SOM-VAE model \citep{DBLP:journals/corr/abs-1806-02199}, inspired by the VQ-VAE architecture \citep{DBLP:journals/corr/abs-1711-00937}
(which itself was later extended by \citet{razavi2019generating}), 
uses an AE to embed the input data points into a latent space and then applies a SOM-based clustering loss on top of this latent representation.
Even though it prominently features a VAE in its name as well as model description, in practice it uses a Dirac $\delta$-distribution and therefore hard assignments of data points to cluster centroids \citep{DBLP:journals/corr/abs-1806-02199}. It also uses a uniform prior over cluster assignments, such that the KL-term is dropped from the loss, thus effectively turning the used model into a standard autoencoder.
Moreover, it employs a Markov model for the temporal dynamics. Both of these design choices yield inferior expressivity compared to our proposed method.
The Deep Embedded SOM, DESOM \citep{Forest2019}, improved the previous model by using a Gaussian neighborhood window with exponential radius decay and by learning the SOM structure in a continuous setting. Both methods extract a topologically interpretable neighborhood structure and yield promising results in visualizing 
state spaces. However, those works did not include empirical comparisons to state-of-the-art deep clustering techniques. Moreover, they do not allow for a probabilistic interpretation of the cluster assignments.
A concise overview of the differences between our proposed models and the related approaches is shown in Table~\ref{table:overview_methods}.

\section{Probabilistic clustering with the DPSOM}
\label{sec:model}

Given a set of data samples $\{ x_i\}_{i=1, \dots, N}$, where $ x_i \in \mathbb{R}^d$, the goal is to partition the data into a set of clusters $\{S_j\}_{j=1, \dots, K}$, while retaining a topological structure over the cluster centroids. 

\begin{figure}[h!]
	\centering
\includegraphics[width=0.7\textwidth]{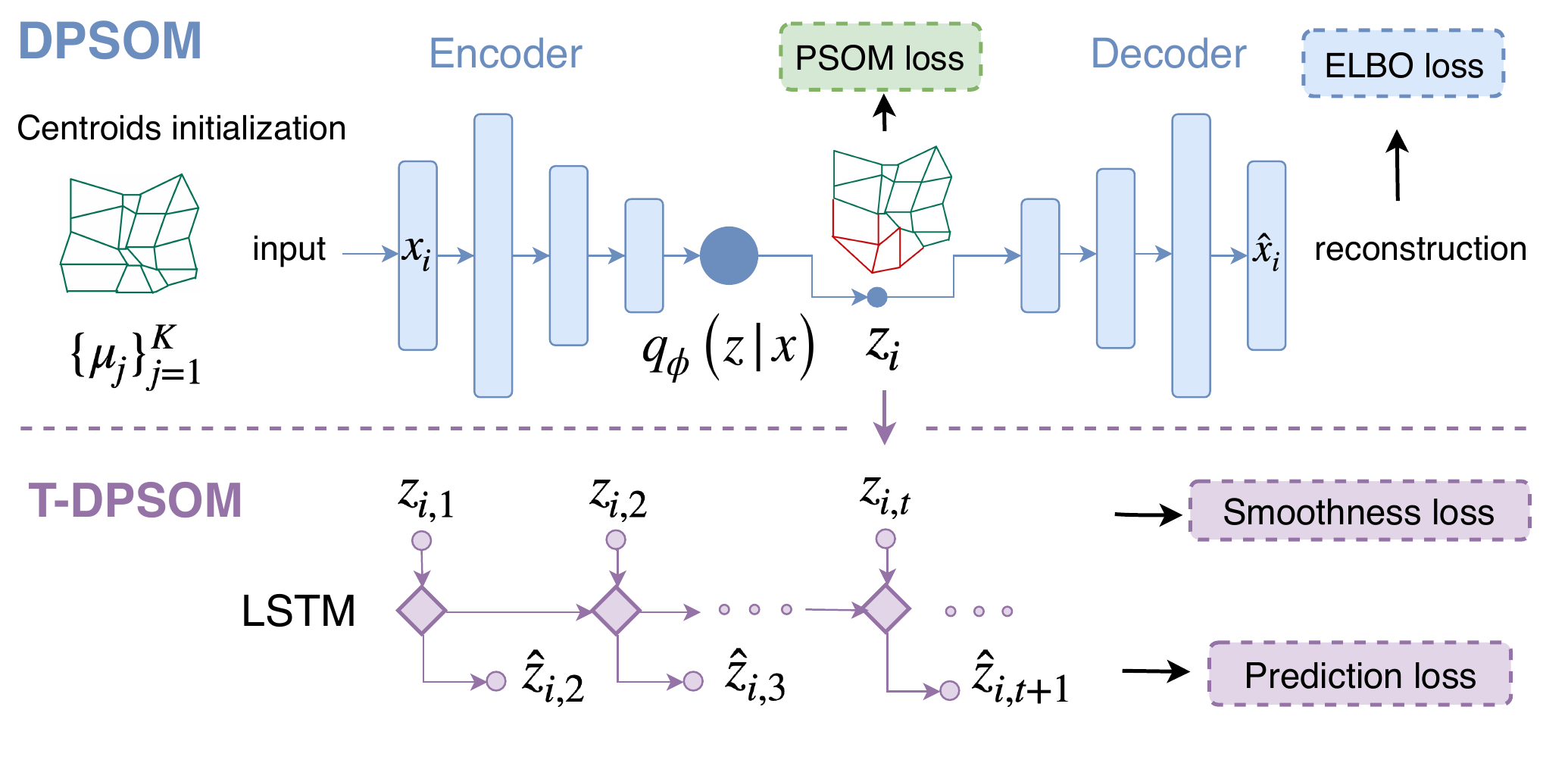}
  \caption{Model architectures of DPSOM and its temporal extension T-DPSOM. A data point $x_i$ is mapped to a continuous embedding $z_i$ using a VAE. In T-DPSOM, the embeddings $z_{i,t} $ for $t=1, \dots, T$ are connected by an LSTM, which predicts the embedding $z_{t+1}$ of the next time step.}
  \label{fig:arch}
\end{figure}

The proposed architecture is presented in Figure \ref{fig:arch}. The input vector $x_i$ is embedded into a 
latent representation $z_i$ using a VAE. This latent vector is then clustered using PSOM, our novel SOM clustering strategy for probabilistic cluster assignments. The VAE and PSOM are trained jointly to learn a latent representation, with the aim of improving the clustering performance. 
To model time series, we propose an architecture extension called T-DPSOM.

\subsection{Background} \label{subsec:som_cah}

A self-organizing map is comprised of $K$ nodes $M = \{ m_j\}_{j=1}^K$ tied by a neighborhood relation, where the node $m_{j}$ corresponds to a centroid $\mu_{j}$ in the input space. 
Given a random initialization of the centroids, the SOM algorithm randomly selects an input $x_i$ and updates both its closest centroid $\mu_{j}$ and its neighbors to move them closer to $x_i$. For a complete description of the SOM algorithm, we
refer to the appendix (Sec.~\ref{sec:self_organizing_maps}).

The Cluster Assignment Hardening (CAH) method has been recently introduced by the DEC model~\citep{DBLP:journals/corr/XieGF15} and
was shown to perform well in the latent space of AEs \citep{DBLP:journals/corr/abs-1801-07648}. Given an embedding 
function $z_i = f(x_i)$, it uses a Student's t-distribution ($S$) as a kernel to measure the similarity between an embedded data point $z_i$, and a centroid $\mu_j$. It improves the cluster purity by forcing the distribution $S$ to approach a target distribution $T$:
\begin{align}
\label{eq:soft_assignments}
     s_{i j}=\frac{\left(1+\left\| z_{i}-\mu_{j}\right\|^{2} / \alpha\right)^{-\frac{\alpha+1}{2}}}{\sum_{j^{\prime}}\left(1+\left\|z_{i}-\mu_{j^{\prime}}\right\|^{2} / \alpha\right)^{-\frac{\alpha+1}{2}}} \; ; \qquad  t_{ij}= \frac{s_{ij}^\kappa / \sum_{i^{\prime}}s_{i^{\prime}j}}{\sum_{j^{\prime}}s_{ij^{\prime}}^\kappa / \sum_{i^{\prime} }s_{i^{\prime}j^{\prime}}}.
\end{align}
By taking the original distribution to the power of $\kappa$ and normalizing it, the target distribution 
puts more emphasis on data points that are assigned a high confidence and thus reduces the entropy of the distribution.
Over the course of training, this lets the distribution approach a discrete cluster assignment (hence ``hardening'').
We follow 
\citet{DBLP:journals/corr/XieGF15} in choosing $\kappa=2$, which leads to larger gradient contributions
of points close to cluster centers, as they show empirically. 
The resulting clustering loss is defined as:
\begin{align}
\label{Lhal}
    \mathcal{L_{\text{CAH}}} = KL(T \Vert S) = \sum_{i=1}^N \sum_{j=1}^K t_{ij} \log{\frac{t_{ij}}{s_{ij}}}.
\end{align}

\subsection{PSOM: Probabilistic SOM clustering} \label{subsec:psom_clustering}

We propose a novel clustering method called Probabilistic SOM (PSOM), which extends the CAH method to include a SOM neighborhood structure over the centroids.
We achieve this by combining (\ref{Lhal}) with a new objective function $\mathcal{L}_{\text{S-SOM}}$ (Soft SOM loss) to get an interpretable representation.
This function maximizes the similarity between each data point and the neighbors of the closest centroids.
Therefore, it acts on soft cluster assignments, but still yields the qualitative behaviour of the SOM algorithm.
The objective is presented in the following.

Given the set of $K$ nodes, $M = \{ m_j\}_{j=1}^K$, we define the neighborhood function as
$N\left(j\right) = \{n_z(j)\}_{z=1}^Z$ for $j \in \{1, \dots, K\}$, where $n_z(j)$ returns the $z$-th neighbor's index of $m_j$.
We require that $\cup_j m_{n_z(j)} = M$, which can for instance be achieved by using a toroid geometry for the map with one neighbor in each direction (the setting used in our experiments).
Each node corresponds to a centroid $\mu_j$ in the latent space.
We use $s_{ij}$, defined in \eqref{eq:soft_assignments}, as the probability that data point $z_i$ belongs to cluster centroid $\mu_j$.
We then define a loss that enforces a SOM-like neighborhood structure over the centroids as:
\begin{align*}
    \mathcal{L}_{\text{S-SOM}} &= -  \frac{1}{N}\sum_{i=1}^N \sum_{j=1}^K s_{ij}\sum_{e \in N(j)} \log{s_{ie}} = \sum_{z=1}^Z -  \frac{1}{N}\sum_{i=1}^N \sum_{j=1}^K s_{ij} \log s_{in_z(j)} \; .
\end{align*}
Intuitively, the objective encourages that if $s_{ij}$ is large, then the $s_{ie}$'s should also be large and vice versa.
Hence, this procedure leads to the same topological neighborhood properties as the classical SOM, while using soft cluster assignments.
Interestingly, this loss can also be seen as a sum of $Z$ cross-entropies between the probability distribution of each centroid and the probability distribution of its respective $z$-th neighbor centroid.
Note that $\sum_j s_{in_z(j) } = 1$ because of the union property in the definition above. The complete PSOM clustering loss is then defined as:
\begin{align}
\label{loss:PSOM}
    \mathcal{L}_{\text{PSOM}} = \mathcal{L_{\text{CAH}}} + \beta \mathcal{L}_{\text{S-SOM}} \; ,
\end{align}
which for $\beta=0$ becomes equivalent to Cluster Assignment Hardening.
We later show empirically (Sec.~\ref{sec:experiments}) that this novel objective does indeed lead to SOM-like behaviour and thus offers a viable way to fit 
self-organizing maps with probabilistic cluster assignments.
Note that the parameter $\beta$ can be chosen freely as a tradeoff between pure clustering performance and interpretability, where increasing $\beta$ improves the smoothness of the learned map (see Fig.~\ref{fig:beta}).

\subsection{DPSOM: VAE for representation learning}

To increase the expressivity of the PSOM, we apply the clustering in the latent space of a deep representation learning model.
In our method, this nonlinear mapping between the input $x_i$ and embedding $z_i$ is realized by a VAE. 
Instead of directly mapping the input $x_i$ to a latent embedding $z_i$, the VAE learns a probability distribution $q_{\phi}(z_i \mid x_i)$ parameterized as a multivariate normal distribution with mean and covariance $(\mu_{\phi}, \Sigma_{\phi}) = f_{\phi}(x_i)$. Similarly, it also learns the probability distribution of the reconstructed output given a sampled latent embedding $p_{\theta}(x_i \mid z_i)$, where $(\mu_{\theta}, \Sigma_{\theta})=f_{\theta}(z_i)$. Both $f_{\phi}$ and $f_{\theta}$ are neural networks, which are respectively called encoder and decoder.
The VAE loss (ELBO) is:
\begin{align}
\label{elbo}
    \mathcal{L}_{\text{VAE}} = \sum_{i=1}^N &\Big[ - \mathbb{E}_{q_{\phi}(z \mid x_i)}(\log{p_{\theta}(x_i \mid z)})
    +  D_{KL}(q_{\phi}(z \mid x_i) \parallel p(z)) \Big] \; , 
\end{align}
where $p(z)$ is an isotropic Gaussian prior over the latent embeddings. The second term can be interpreted as a form of regularization, which encourages the latent space to be compact. %
For each data point $x_i$, the latent embedding $z_i$ is sampled from $q_{\phi}(z \mid x_i)$.
Adding the VAE loss to the PSOM loss from the previous subsection, we get the overall loss function of the DPSOM:
\begin{align}
\label{loss_DPSOM}
    \mathcal{L}_{\text{DPSOM}} =\gamma  \mathcal{L}_{\text{CAH}} + \beta  \mathcal{L}_{\text{S-SOM}} + \mathcal{L}_{\text{VAE}} \; ,
\end{align}
where $\gamma$ regulates the tradeoff between reconstruction and clustering performances while $\beta$ is as above. For an in-depth discussion of the choice of these parameters and of our model's robustness to different parameter configurations, we refer to the Appendix \ref{subsec:HP}.
To the best of our knowledge, no previous SOM method attempted to use a VAE to embed the inputs into a latent space. 
Yet, the VAE is a natural choice, since the compactness of the representations encouraged by the Gaussian prior fits the neighborhood assumptions of the SOM algorithm (see Appendix~\ref{sec:self_organizing_maps}).

\subsection{T-DPSOM: Extension to time series data}

To extend our proposed model to time series data, we add a temporal component to the architecture, yielding the Temporal DPSOM (T-DPSOM).
Given a set of $N$ time series of length $T$, $\{ x_{i,t}\}_{i=1, \dots, N; t=1,\dots, T}$, the goal is to learn interpretable 
trajectories on the SOM grid. To do so, the DPSOM could be used directly but it would treat each time step $t$ of the time series 
independently. To exploit temporal information and enforce smoothness in the 
trajectories, we design an additional loss term, which is similar to the smoothness loss in the SOM-VAE \citep{DBLP:journals/corr/abs-1806-02199}, but is able to act on soft assignments:
\begin{align}
    \mathcal{L}_{\text{smooth}} =  - \frac{1}{N T}\sum_{i=1}^N \sum_{t=1}^T u_{i_{t},i_{t+1}} \; ,
\end{align}
where $u_{i_{t},i_{t+1}}=g(z_{i,t},z_{i,t+1})$ is the similarity between $z_{i,t}$ and $z_{i,t+1}$ using a Student's t-distribution and $z_{i,t}$ refers to the embedding of time series $x_i$ at time index $t$. It maximizes the similarity between latent embeddings of adjacent time steps, such that large jumps in the latent state between time points are discouraged.
This is motivated by the intuition that the true underlying factors of variation in real-world applications usually vary smoothly over time and can be seen as being similar to a Kalman filter prior \citep{krishnan2016deep}.

One of the main goals in time series modeling is to predict future data points, or alternatively, future embeddings.
This can be achieved by adding a long short-term memory network (LSTM) \citep{hochreiter1997long} over the latent embeddings of the time series,
as shown in Fig \ref{fig:arch}. Each cell of the LSTM takes as input the latent embedding $z_t$ at time step $t$, and predicts a probability distribution over the next latent embedding, $p_{\omega}(z_{t+1}\mid z_t)$. We parameterize this distribution as a multivariate Gaussian distribution where the mean and variance are learnt by the LSTM. The prediction loss is the log-likelihood between the learned distribution and a sample of the next embedding $z_{t+1}$:
\begin{align}
    \mathcal{L}_{\text{pred}} =  - \sum_{i=1}^N \sum_{t=1}^T \log{p_{\omega}(z_{t+1}\mid z_t)} \; .
\end{align}
The final loss of the T-DPSOM, which is trainable in a fully end-to-end fashion, is
\begin{align}
     \mathcal{L}_{\text{T-DPSOM}} = \mathcal{L}_{\text{DPSOM}}+ \mathcal{L}_{\text{smooth}}+ \mathcal{L}_{\text{pred}} \; .
\end{align}
This combined objective encourages the learned representations and clusters to preserve similarity between inputs in its topological structure (through the first term), while also preserving smoothness over time (through the second term), and learning representations that are informative about the future of the trajectory (through the third term).
It therefore ensures usefulness of the representations for clustering, time series visualization, and time series forecasting.
In the following, we will separately evaluate these three use cases empirically.
\section{Experiments}
\label{sec:experiments}

Firstly, we evaluate the DPSOM and compare its clustering performance to a wide range of state-of-the-art deep and SOM-based clustering methods, on 
MNIST \citep{726791} and Fashion-MNIST \citep{DBLP:journals/corr/abs-1708-07747} data.
We then present extensive evidence of the performance of the T-DPSOM on medical time series from 
the eICU data set \citep{pollard2018eicu} on several relevant tasks. Moreover, we discuss how the spatial coherence of the clusterings depends on the used objective functions, for
both MNIST and the medical data. Lastly, we illustrate how the probabilistic assignments of T-DPSOM enable interpretable
visualizations of medical time series with uncertainty estimation. The datasets are described in detail in Appendix \ref{subsec:datasets}.%

\begin{figure*}[!b]
	\centering
	\subfloat[$\beta=0$\label{fig:b1}]{\includegraphics[height=0.15\linewidth]{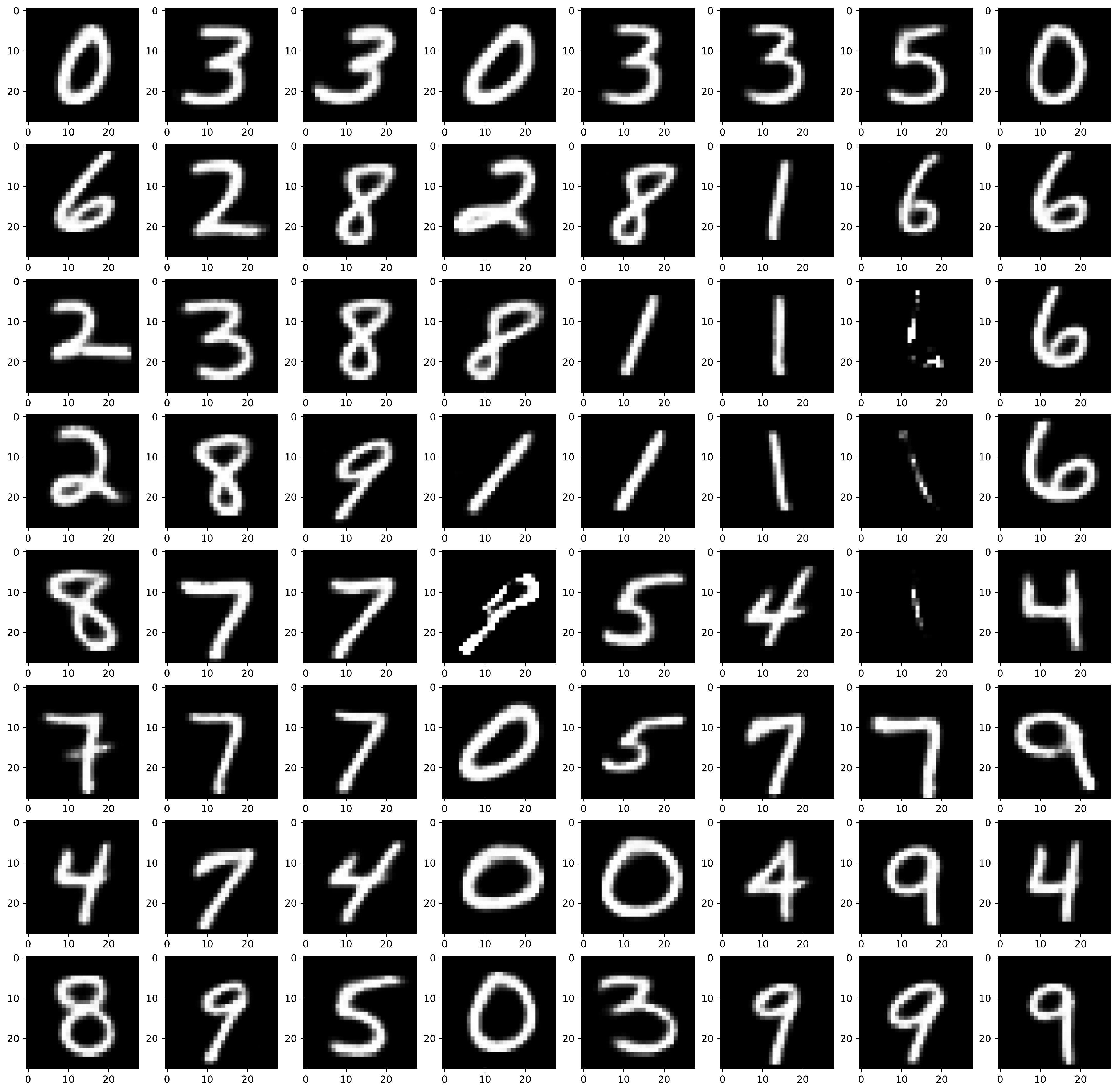}}
	\hfill
	\subfloat[$\beta=1$\label{fig:b2}]{\includegraphics[height=0.15\linewidth]{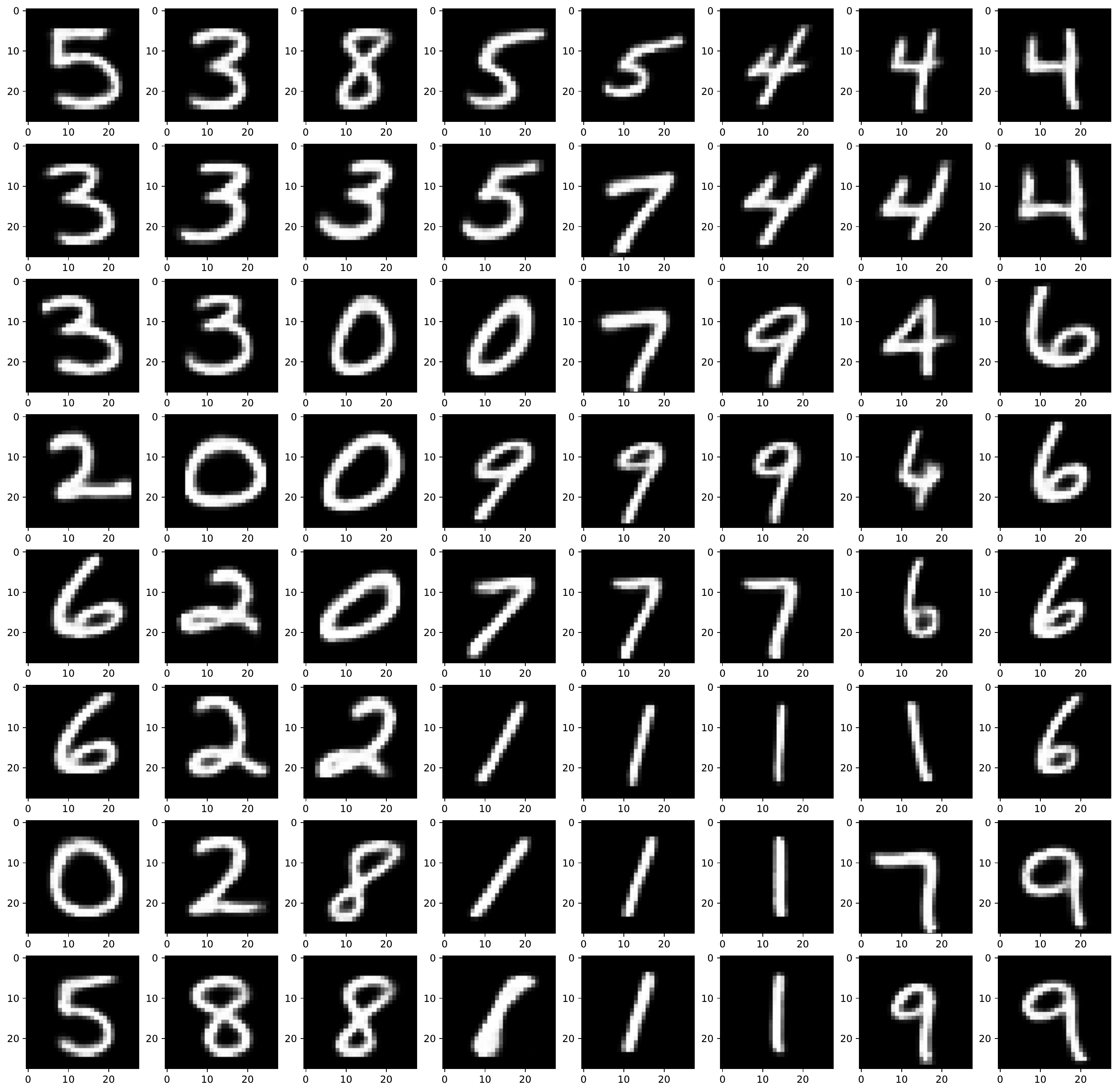}}
	\hfill
	\subfloat[$\text{MI}=0.483$]{\includegraphics[height=0.15\linewidth]{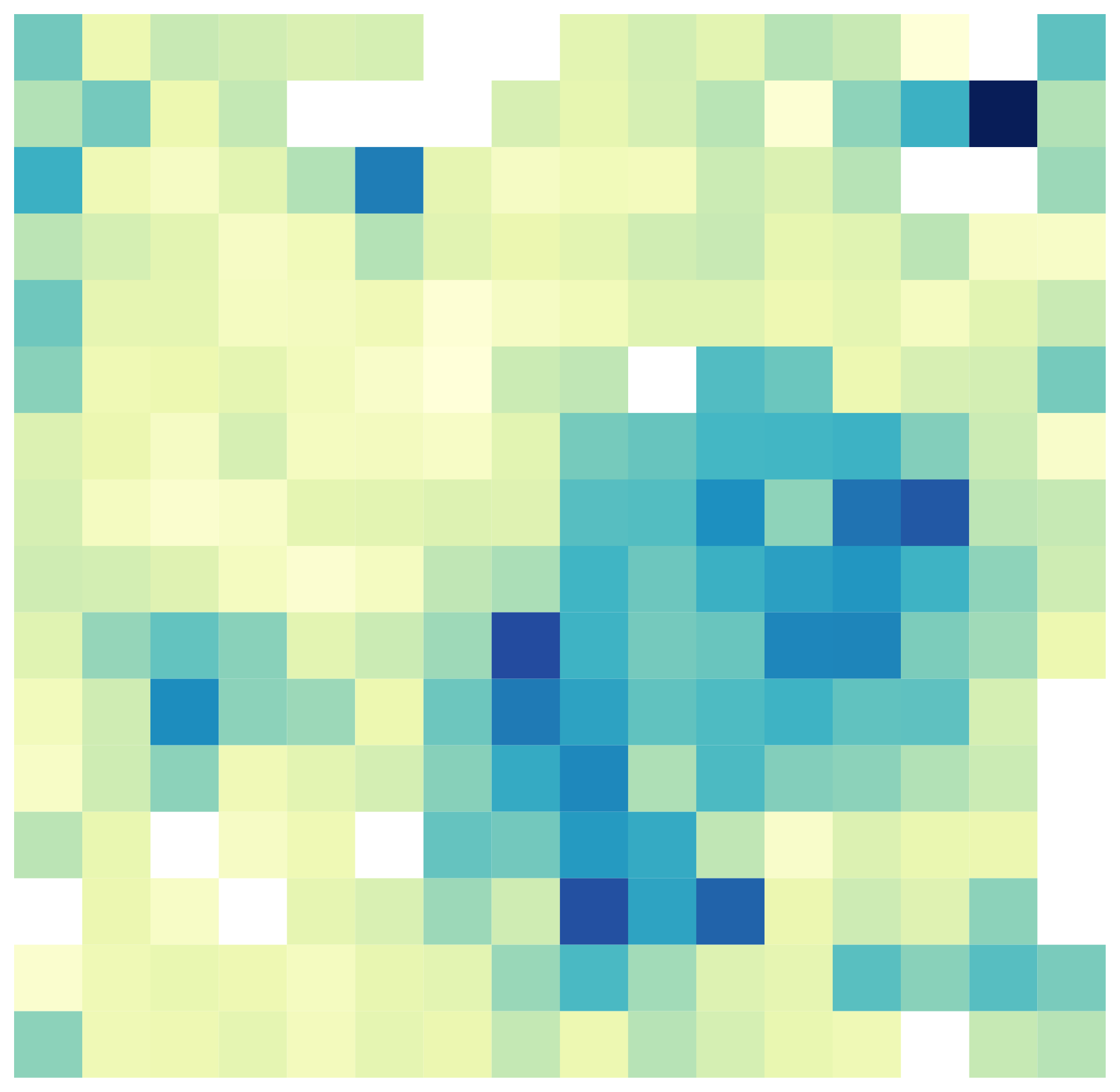}}
	\hfill
	\subfloat[$\text{MI}=0.626$]{\includegraphics[height=0.15\linewidth]{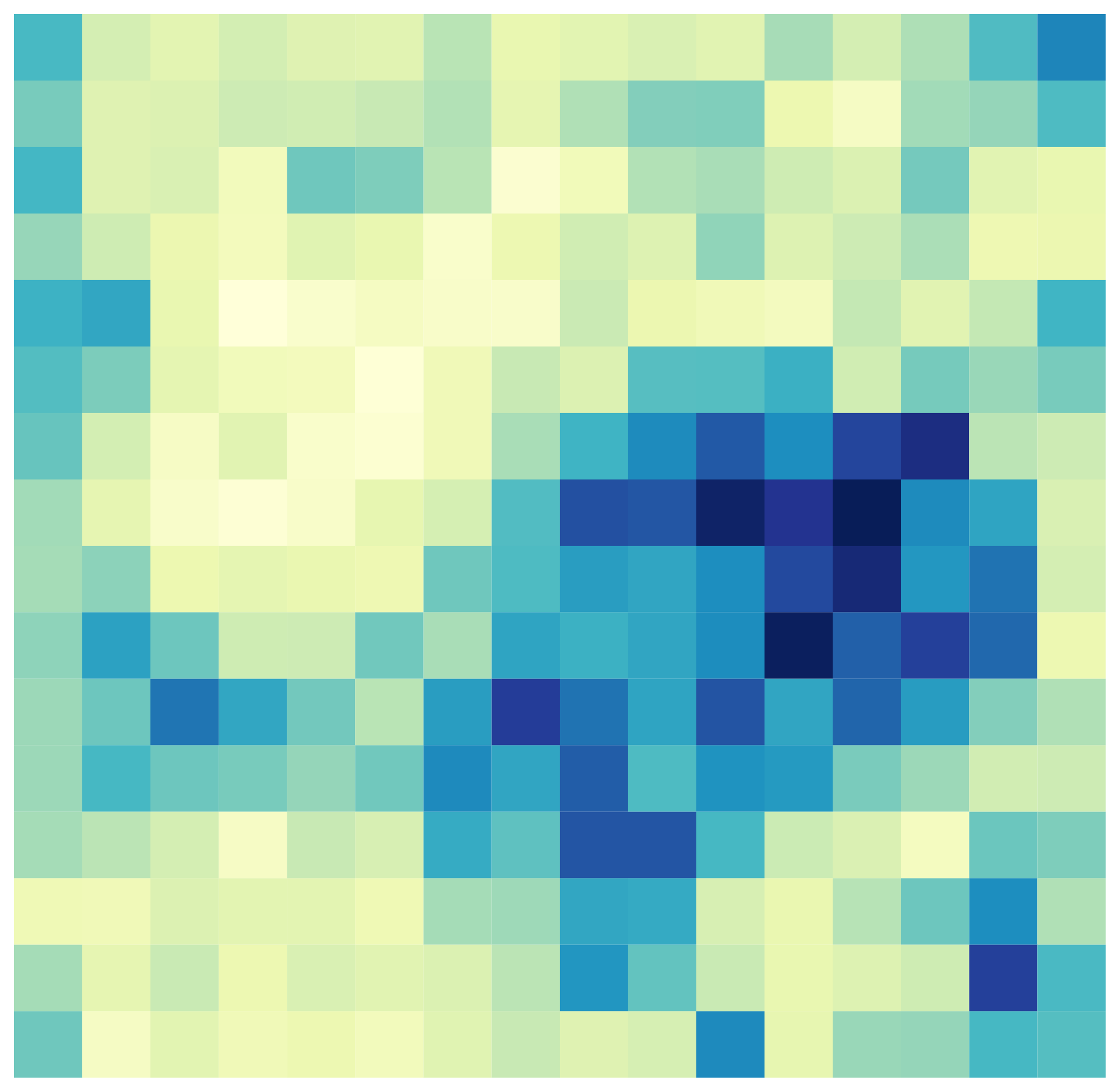}}
	\hfill
	\subfloat[$\text{MI}=0.749$]{\includegraphics[height=0.15\linewidth]{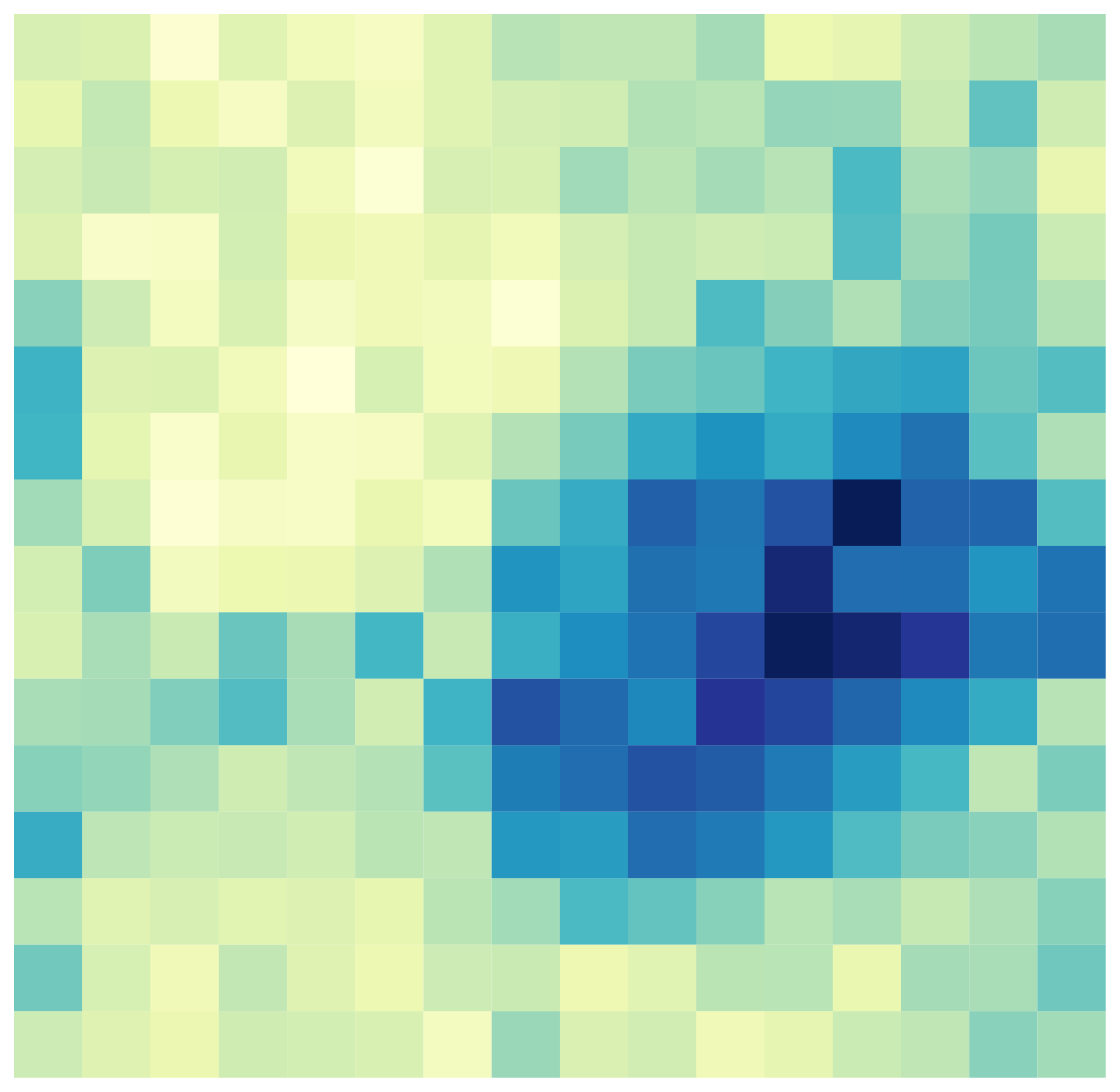}}
	\hfill
	\subfloat[$\text{ MI}=0.764$]{\includegraphics[height=0.15\linewidth]{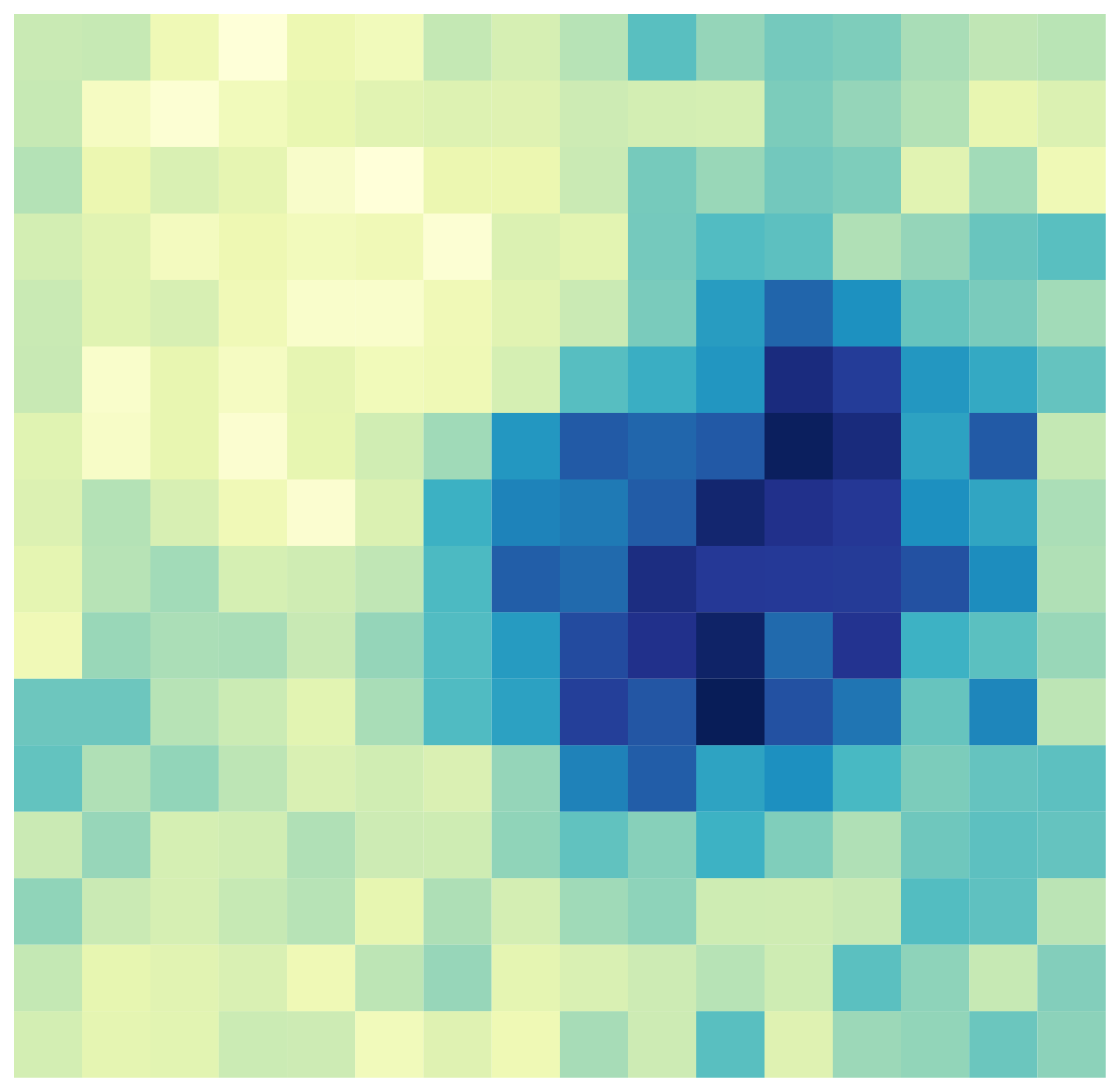}}
	\hfill
	\subfloat{\includegraphics[trim=0 7 0 0, height=0.15\linewidth]{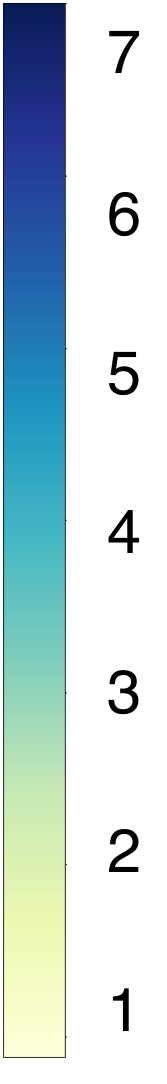}}
	\hfill
	\caption{(a)-(b) Visualizations of the SOM grid reconstructions obtained by training DPSOMs with and without S-SOM loss on MNIST. (c)-(f) Visualizations of the SOM grid heat-maps obtained by training T-DPSOMs with $\beta= 0$, $\beta= 10$, $\beta= 50$ and  $\beta= 100$ respectively. We see that increasing $\beta$ increases the correlation between neighboring clusters, which is also shown quantitatively using Moran's index (MI). For all experiments we used the same random initialization.}
	\label{fig:beta}
\end{figure*}

\paragraph{Image clustering.}
For the clustering experiments on static data, we used two different categories of baselines. The first category contains clustering methods that do not provide any interpretable discrete latent representation. Those include k-means, the DEC model \citep{DBLP:journals/corr/XieGF15}, as well as its improved version IDEC \citep{ijcai2017-243}.
We also include the VQ-VAE \citep{DBLP:journals/corr/abs-1711-00937}, which formed the basis for the SOM-VAE model \citep{DBLP:journals/corr/abs-1806-02199}.
In the second category, we include state-of-the-art clustering methods based on SOMs.
Here, we used a standard SOM %
, AE+SOM, an architecture composed of an AE and a SOM applied on top of the latent representation
(trained sequentially), SOM-VAE \citep{DBLP:journals/corr/abs-1806-02199}, and DESOM \citep{Forest2019}. 
\newline
To implement our model we focused on retaining a fair comparison with the baselines, hence both the AE of the baselines and the VAE of our model use the same standard network structure.
The number of clusters is set to $64$ (arranged in an $8 \times 8$ grid for the SOM methods), $\gamma$ is set to $20$ (Equation \ref{loss_DPSOM}) and $\beta$ is chosen in an unsupervised way such that the CAH loss and the S-SOM loss have similar magnitude ($0.25$ on MNIST and $0.4$ on fMNIST).
For details on the architecture, as well as its implementation we refer to the Appendix \ref{subsec:implementation}.
\newline
Table \ref{table:1} shows the clustering performance of DPSOM on MNIST and Fashion-MNIST data, compared 
with the baselines. Purity and Normalized Mutual Information are used as evaluation metrics. 
We observe that our proposed model outperforms
the baselines in terms of both metrics on both data sets. 
Interestingly, the DPSOM not only improves interpretability through the use of the latent PSOM, but also increases the performance compared to DEC/IDEC. 
We tested the robustness of our model by randomly drawing the values of both hyperparameters 
from $\beta \in [0,1], \gamma \in [10,30]$ and we observed similar clustering performance (see Appendix \ref{subsec:HP}). Additionally, DPSOM outperforms 
its main competitor IDEC also using other choices for the number of clusters (see Appendix \ref{subsec:num_clusters}). 
\newline
Finally, we performed two ablation studies to investigate the effect of different components of the model.
In the first ablation, we removed the S-SOM loss (setting $\beta = 0$) and we noticed similar performances compared to DPSOM. The main motivation for the addition of the S-SOM loss is to improve interpretability in the latent space (see Fig.~\ref{fig:beta}) and this is achieved without compromising the clustering performances, which is a quite remarkable result.
We also exchanged the VAE with an AE and we observe a significant decrease in performance.
This shows that both components are indeed integral for our model to perform its intended function.

\begin{table*}[ht!]
\caption{Clustering performance of DPSOM using 64 clusters arranged in a $8\times8$ SOM map, compared with baselines. %
	Means and standard errors are computed across 
         10 runs with different random model initializations.
         *The DESOM results are taken from \citep{Forest2019} which did not provide errors. %
        }
\label{table:1}
\vspace{-0.2cm}
\begin{center}
\resizebox{\linewidth}{!}{
\begin{tabular}{@{}lcccccc@{}}\toprule
& \multicolumn{2}{c}{MNIST} & & \multicolumn{2}{c}{fMNIST}  \\ 
 \cmidrule{2-3} \cmidrule{5-6}   
& $pur$ & $nmi$ && $pur$ & $nmi$ 
\\ \midrule
K-means & $0.845 \pm 0.001$ & $0.581 \pm 0.001 $ && $0.716 \pm 0.001$ & $0.514 \pm 0.000$  \\ 
 VQ-VAE & $0.515 \pm 0.005$ & $0.354 \pm 0.003$ && $0.594 \pm 0.003$ & $0.468 \pm 0.001$ \\ %
 DEC & $0.944 \pm 0.002$ & $0.682 \pm 0.001$ && $0.758 \pm 0.002$  & $\mathbf{0.562 \pm 0.001}$  \\ 
 IDEC & $0.950 \pm 0.001$ & $0.681 \pm 0.001$  && -   & -  \\ 
 \midrule
SOM & $0.701 \pm 0.005$ & $0.539 \pm 0.002$ && $0.667 \pm 0.003$ & $0.525 \pm 0.001$ \\
 AE+SOM & $ 0.874 \pm 0.004$ & $0.646 \pm 0.001$ && $0.706 \pm 0.002$ & $0.543 \pm 0.001$  \\ 
 SOM-VAE & $0.868 \pm 0.004 $ & $0.595 \pm 0.004$ && $0.739 \pm 0.005$ & $0.520 \pm 0.003$  \\ 
 DESOM* & $0.939  \pm \; N/A$ & $0.657 \pm \; N/A$ && $0.752 \pm \; N/A$ & $0.538 \pm \; N/A$  \\ 
 DPSOM (ours) & $\mathbf{0.968 \pm 0.001} $& $\mathbf{0.701 \pm 0.001}$  && $\mathbf{0.779 \pm 0.003}$ & $\mathbf{0.562 \pm 0.001} $\\
  \midrule
  DPSOM $\backslash$ PSOM (ablation) & $0.965 \pm 0.001 $ & $\mathbf{0.701 \pm 0.001}$  && $ 0.770 \pm 0.002$ & $\mathbf{0.563 \pm 0.001}$  \\ 
  DPSOM $\backslash$ VAE (ablation) & $ 0.813 \pm 0.004 $&$ 0.561 \pm 0.002 $ && $ 0.730 \pm 0.006 $&$ 0.530 \pm 0.003 $ \\
 \bottomrule
\end{tabular}
}
\end{center}
\end{table*}

\paragraph{Clustering and forecasting of time series states.}
The clustering performance of our proposed models was evaluated on the eICU data set \citep{pollard2018eicu}, which is comprised of multivariate medical time series from the intensive care unit (ICU). We compare them against the SOM-VAE (the only deep clustering method among the baselines that is designed for temporal data), k-means and an HMM. 
Table \ref{table:2} shows the cluster enrichment in terms of NMI for four different labels, the current (APACHE-0) and worst future (APACHE-6/12/24) physiology scores in the next $6,12,24$ hours, respectively. The T-DPSOM clearly achieves superior 
clustering performance compared to the baselines.
\newline
To quantify the performance of T-DPSOM in predicting future trajectories, we predict the final six latent embeddings of each time series, conditioned on all previous time steps. For each predicted embedding, we reconstruct the input using the decoder of the VAE. Finally, we measure the mean squared error (MSE) between the original inputs and the reconstructed inputs for the last six hours of the ICU stay. As baselines, we use an LSTM, an HMM, and the SOM-VAE. The LSTM predicts the future time steps directly in the input space. The HMM predicts by sampling future time steps using its learned transition and emission matrices, starting from the most likely current state.
The results (Table \ref{table:3}) indicate that the joint training of clustering and prediction used by T-DPSOM clearly outperforms the baselines. Training times of the 
T-DPSOM model are reported in Appendix \ref{sec:comp-issues}.

\begin{table*}[ht!]
\caption{Mean NMI and standard error of cluster enrichment for current/future APACHE physiology scores, 
         using a 16 $\times$ 16 SOM map, across 10 runs with different random model initializations. }%
\label{table:2}
\vspace{-0.2cm}
\begin{center}
\resizebox{\linewidth}{!}{
\begin{tabular}{@{}lcccc@{}}\toprule
Model & APACHE-24 & APACHE-12 & APACHE-6 & APACHE-0 
\\ \midrule
 K-means & $0.0967 \pm 0.0034$ & $0.0862 \pm 0.0034$ & $0.0837 \pm 0.0031$ &  $ 0.0905 \pm  0.0031$ \\ 
 SOM-VAE & $ 0.0824 \pm 0.0008 $&$0.0758 \pm 0.0007$ & $0.0743 \pm 0.0007$  & $0.0803 \pm 0.0008$\\
 HMM & $ 0.0533 \pm 0.0007 $&$0.0463 \pm 0.0007$ & $0.0427 \pm 0.0007$  & $0.0419 \pm 0.0008$\\ %
 DPSOM  &$0.0919 \pm 0.0037$ & $0.0843  \pm 0.0031 $ & $ 0.0816  \pm 0.0029$ & $0.0875  \pm  0.0033$ \\
 T-DPSOM  & $\mathbf{0.1115 \pm 0.0006}$&$\mathbf{0.10220 \pm 0.0005}$  & $\mathbf{0.0989 \pm 0.0004}$ & $\mathbf{0.1065 \pm 0.0005}$\\
\bottomrule
\end{tabular}
}
\end{center}
\vspace{-0.2cm}
\end{table*}
\FloatBarrier

\begin{table}[ht!]
\caption{MSE for predicting the time series of the last 6 hours before ICU dispatch, 
         given the prior time series, across 10 runs with different random model initializations.}
\label{table:3}
\begin{center}
\begin{tabular}{@{}lcccc@{}}\toprule
Model & LSTM & HMM & SOM-VAE & T-DPSOM 
\\ \midrule
MSE & $0.0113\pm 0.0002$ & $0.0146 \pm 0.0001$ & $0.0081 \pm 0.0001$ & $\mathbf{0.0049 \pm 0.0001}$\\ 
\bottomrule
\end{tabular}
\end{center}
\vspace{-0.3cm}
\end{table}

\paragraph{Spatial coherence through S-SOM loss.}
\label{subsec:interpretability}
\looseness-1

The main objective of the PSOM is to enforce a SOM-like structure between the cluster centroids.
Thus, we achieve an interpretable 2-D representation of the data in which neighboring centroids should exhibit similar characteristics.
To illustrate the topological structure in the latent space, we present the reconstructions of the DPSOM centroids, arranged 
in an $8 \times 8$ grid, on MNIST data in Figure~\ref{fig:b1}/\ref{fig:b2} with and without the S-SOM loss. 
We see that similar digits are more tightly clustered in neighbouring centroids in Figure~\ref{fig:b2}.
To assess this property quantitatively, we use Moran's index (MI) \citep{10.1093/biomet/37.1-2.17} as a measure of spatial correlation among clusters.
Moran's index is defined as:
\begin{align*}
I=\frac{N}{W} \frac{\sum_{i} \sum_{j} w_{i j}\left(y_{i}-\bar{y}\right)\left(y_{j}-\bar{y}\right)}{\sum_{i}\left(y_{i}-\bar{y}\right)^{2}} \;,
\end{align*}
where $N$ is the number of clusters indexed by $i$ and $j$, $y_i$ is the variable of interest, $\bar{y}$ is the mean of $y$, and $w_{i,j}$ is a matrix of spatial weights.
We define 
$
w_{i,j} = \exp(-d_{\text{SOM}}(i,j))
$,
where $d_{\text{SOM}}(i,j)$ is the distance between the nodes $m_i$ and $m_j$ in the SOM.
We use the eICU data set with the mean APACHE score as the cell label to express 
similarities between patient states. We compute Moran's index for different values of the parameter $\beta$ in Equation \ref{loss:PSOM}.
To qualitatively show the effect of the S-SOM loss, we include heatmaps that show the enrichment of cells for the current APACHE physiology score. 
We see in Figure~\ref{fig:beta}(c)-(f) that increasing the $\beta$ coefficient, and thus the relative weight of the S-SOM loss, increases the correlations between neighboring clusters, both visually and quantitatively.
We can thus conclude that the S-SOM loss does indeed encourage the latent representations to assume a SOM-like structure, and that the spatial coherence of the clustering can be controlled via $\beta$.

\paragraph{Interpretable visualization of time series with uncertainty estimation.}
\label{subsec:soft_ass}

\begin{figure*}[ht]
	\centering
	\hfill
	\subfloat{\includegraphics[height=0.18\linewidth]{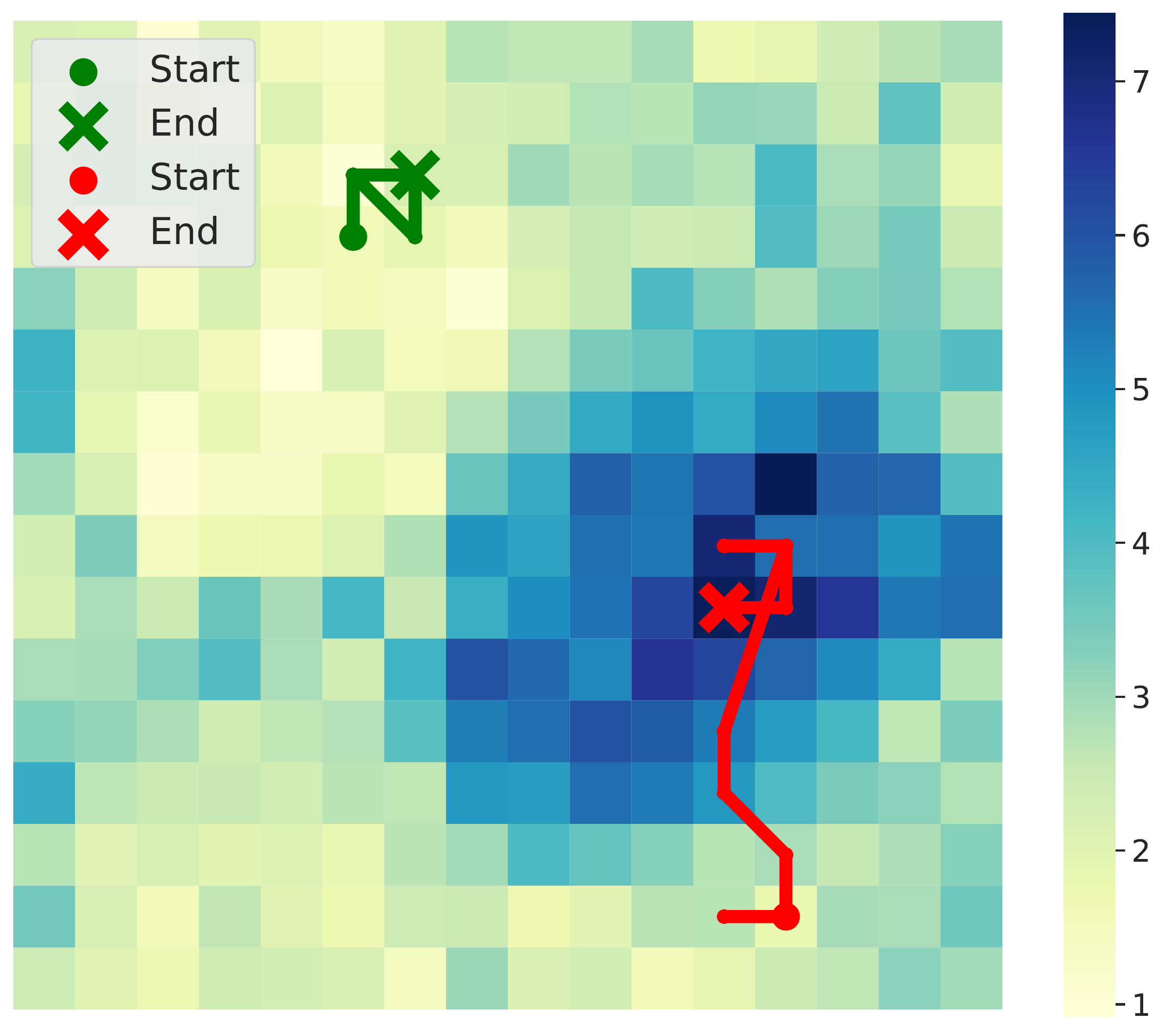}}
	\hfill
	\subfloat{\includegraphics[height=0.18\linewidth]{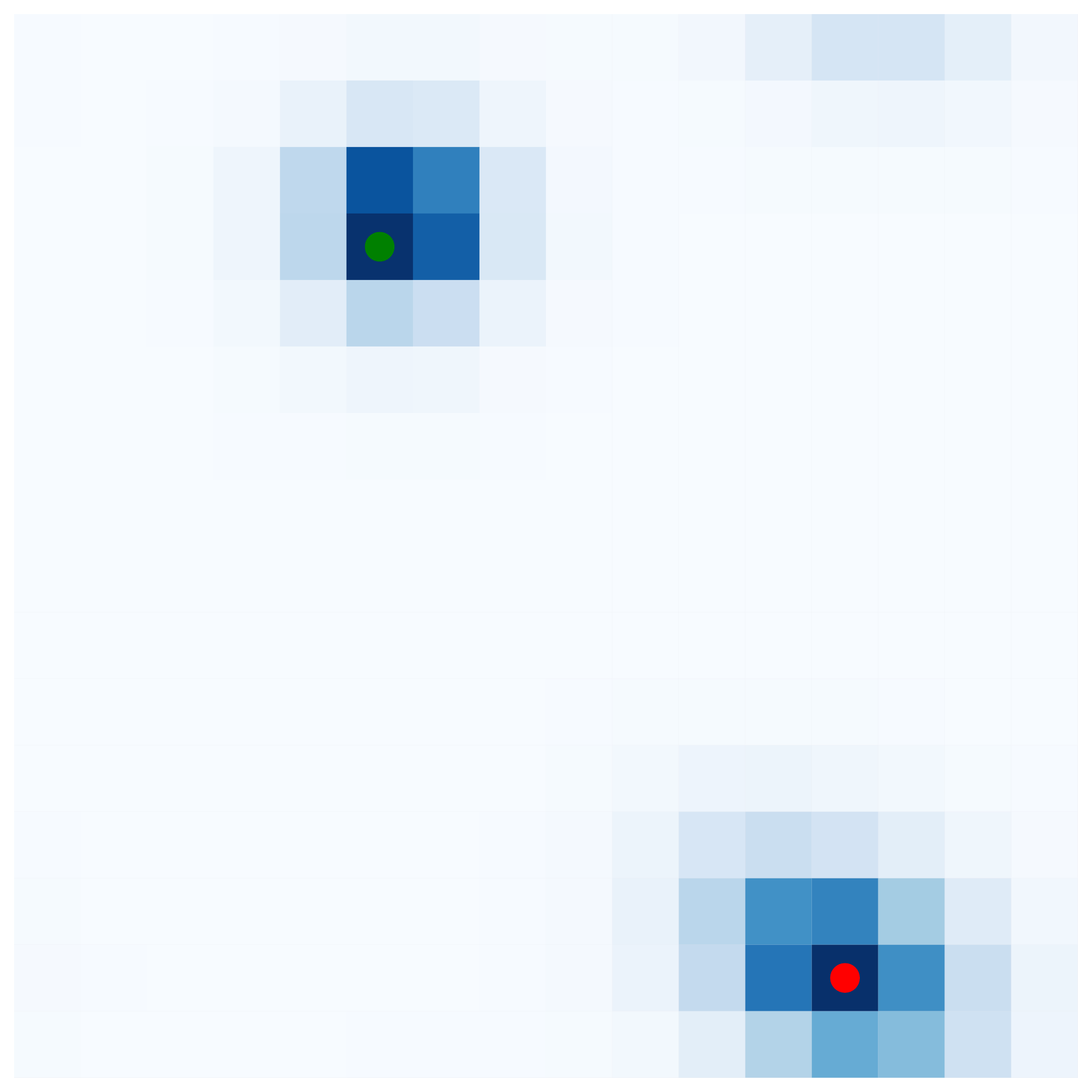}}
	\hfill
	\subfloat{\includegraphics[height=0.18\linewidth]{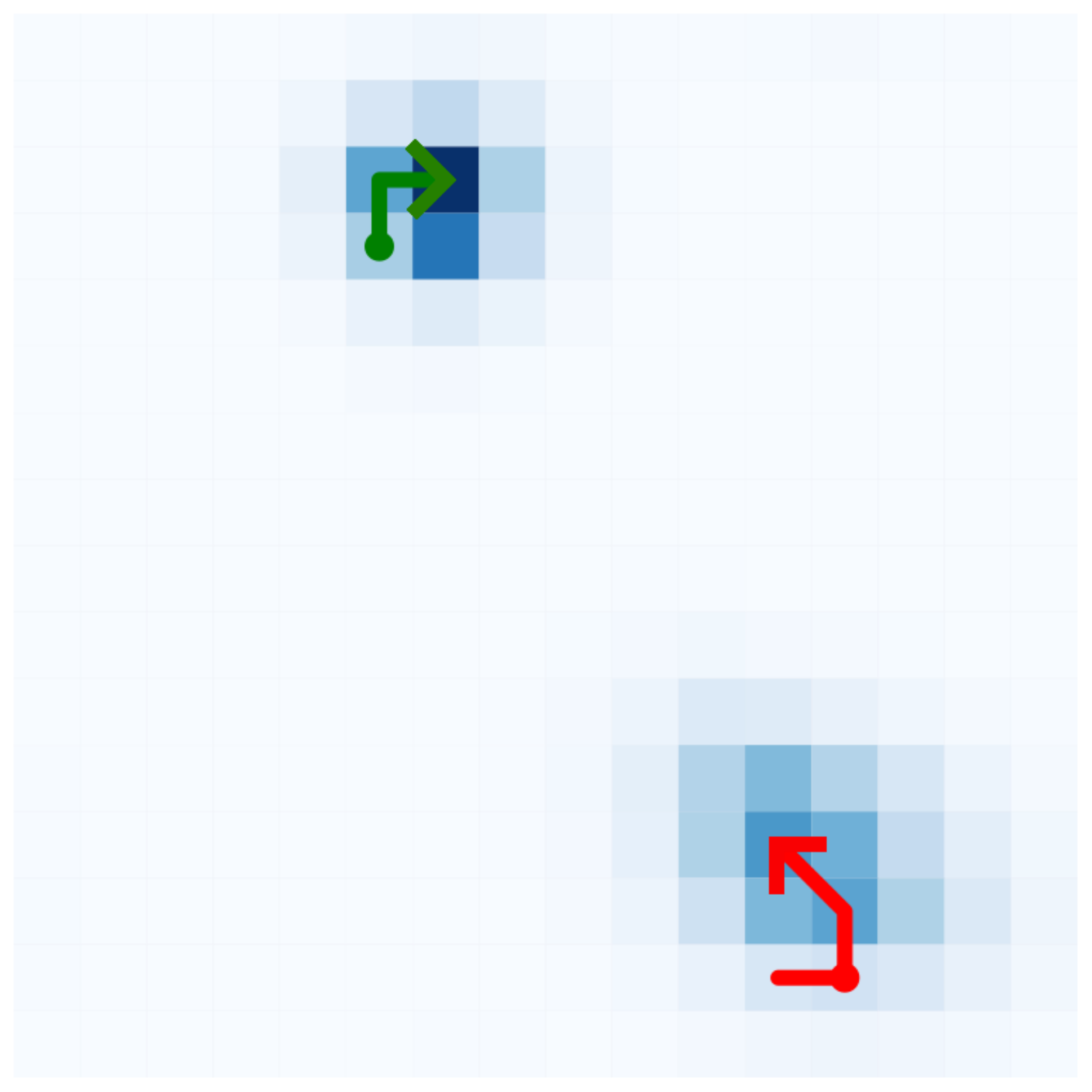}}
	\hfill
	\subfloat{\includegraphics[height=0.18\linewidth]{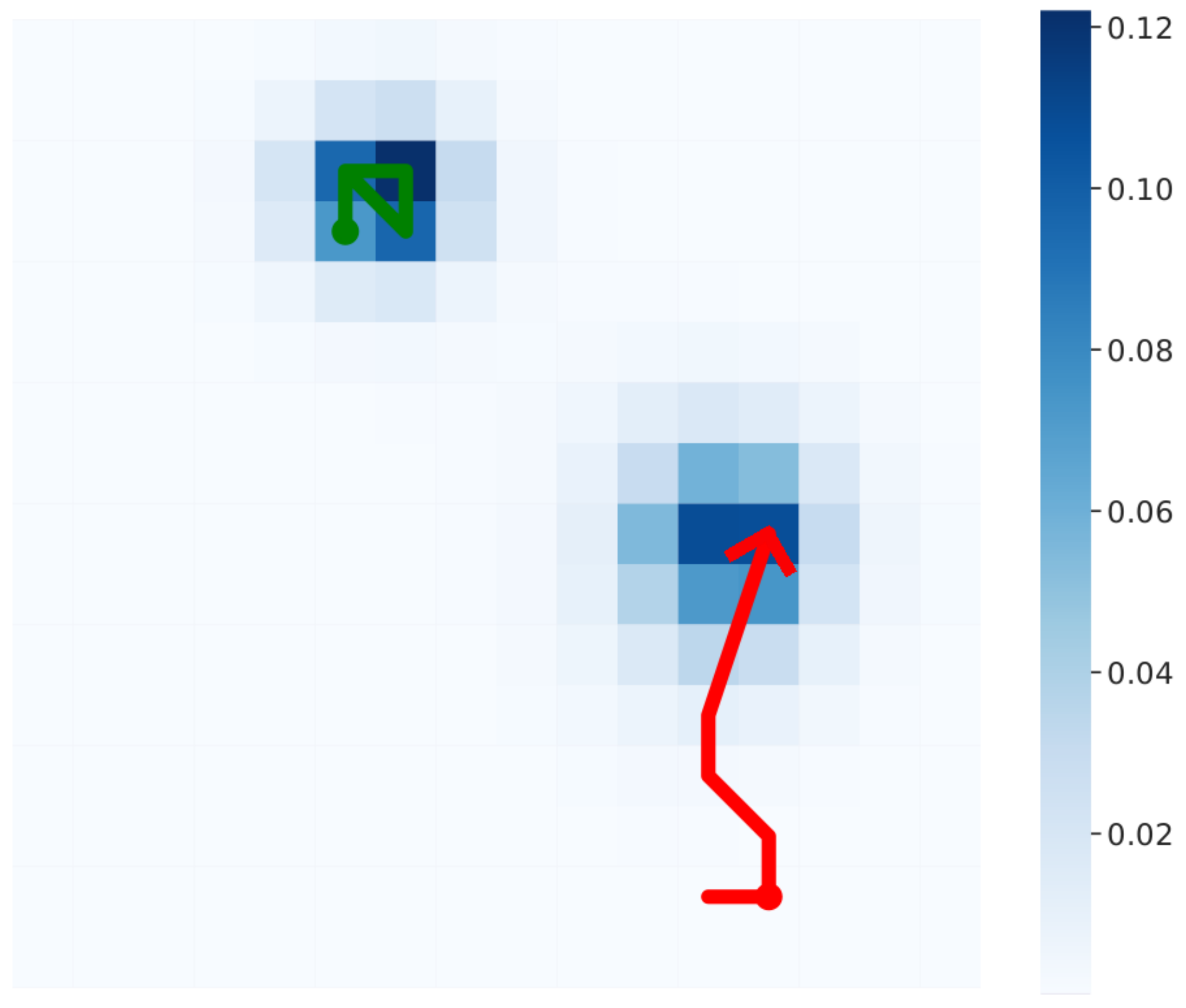}}
	\hfill
	\hfill
	\caption{Illustration of two example patient trajectories in the SOM grid of T-DPSOM. One patient died (red), while the other was dispatched alive from the ICU (green). Superimposed is a heatmap which displays mean APACHE score of all time points assigned to each cluster. We observe qualitative differences in the trajectories of the dying and the surviving patient. For each time series we also show the assigned probabilities to the discrete patient health states using a blue color shading.}
	\label{fig:prob}
\end{figure*}

On the ICU time series data, we show example trajectories for one patient dying at the end of the ICU stay, as well as a control patient who is dispatched alive from the ICU. 
We observe that the trajectories are located in different parts of the SOM grid, and that their directions of movement fit the intuition when combined with the average physiology scores of each cluster (Fig.~\ref{fig:prob}).
One of the advantages of the T-DPSOM over the SOM-VAE algorithm is the use of 
soft assignments of data points to clusters, which results in the ability to quantify 
uncertainty in the clustering. For interpreting health states in the ICU, this property is very important \cite{hudson2010uncertainty}.
In Figure~\ref{fig:prob}, additionally to the patient trajectories, we show the probability distributions over cluster assignments at different time steps. Our model yields a soft centroid-based probability distribution which evolves over time 
and which allows estimation of likely discrete health states at any given point in time. For each time step, the distribution 
of probabilities is plotted using a blue color shading, whereas the overall trajectory is plotted using a solid line. 
\newline
We see that the assigned probabilities fit well to the intuition that neighboring clusters should be harder to distinguish than more separated ones.
Moreover, neighboring clusters with larger assigned probability can sometimes forebode the movement direction of the trajectory, suggesting that the combination of SOM-loss and temporal losses leads to a representation that is smooth in space as well as in time.
Further results, including a more quantitative evaluation using randomly sampled trajectories and cluster enrichment for future mortality, can be found
in the Appendix \ref{subsec:icu_uncertainty}, \ref{subsec:state_representation_icu}. 
\section{Conclusion}
\label{sec:conclusion}
\looseness -1
We presented two novel methods for interpretable unsupervised clustering on static and temporal data, 
DPSOM and T-DPSOM. Both models make use of a VAE and a novel probabilistic clustering method, PSOM, that extends the classical SOM algorithm to include 
centroid-based probability distributions. They achieve superior clustering performance compared to state-of-the-art deep clustering 
baselines on benchmark data sets and medical time series. The use of probabilistic assignments of data points to clusters, and the use of a VAE for feature extraction, instead of an AE as used in previous methods, results in an interpretable model that can quantify uncertainty in the clustering as well as in its predictions of future time series states.

\subsection*{Acknowledgments}

This project was supported by the grant \#2017-110 of the Strategic Focus Area ``Personalized Health and Related Technologies (PHRT)'' of the ETH Domain. 
VF, MH are partially supported by ETH core funding (to GR). MH is supported by the Grant No. 205321\_176005 of the Swiss National 
Science Foundation (to GR).
VF is supported by a PhD fellowship from the Swiss Data Science Center.
We thank Natalia Marciniak for her administrative efforts. 

\bibliography{bibliography/references}

\begin{thebibliography}{39}
\providecommand{\natexlab}[1]{#1}
\providecommand{\url}[1]{\texttt{#1}}
\expandafter\ifx\csname urlstyle\endcsname\relax
  \providecommand{\doi}[1]{doi: #1}\else
  \providecommand{\doi}{doi: \begingroup \urlstyle{rm}\Url}\fi

\bibitem[Aljalbout et~al.(2018)Aljalbout, Golkov, Siddiqui, and
  Cremers]{DBLP:journals/corr/abs-1801-07648}
Elie Aljalbout, Vladimir Golkov, Yawar Siddiqui, and Daniel Cremers.
\newblock Clustering with deep learning: Taxonomy and new methods.
\newblock \emph{CoRR}, abs/1801.07648, 2018.
\newblock URL \url{http://arxiv.org/abs/1801.07648}.

\bibitem[Bishop(2006)]{Bishop:2006:PRM:1162264}
Christopher~M. Bishop.
\newblock \emph{Pattern Recognition and Machine Learning (Information Science
  and Statistics)}.
\newblock Springer-Verlag, Berlin, Heidelberg, 2006.
\newblock ISBN 0387310738.

\bibitem[Bishop et~al.(1998)Bishop, Svens{\'e}n, and Williams]{bishop1998gtm}
Christopher~M Bishop, Markus Svens{\'e}n, and Christopher~KI Williams.
\newblock Gtm: The generative topographic mapping.
\newblock \emph{Neural computation}, 10\penalty0 (1):\penalty0 215--234, 1998.

\bibitem[Chappell and Taylor(1993)]{Chappell:1993:TKM:154879.154890}
Geoffrey~J. Chappell and John~G. Taylor.
\newblock The temporal kohonen map.
\newblock \emph{Neural Netw.}, 6\penalty0 (3):\penalty0 441--445, March 1993.
\newblock ISSN 0893-6080.
\newblock \doi{10.1016/0893-6080(93)90011-K}.
\newblock URL \url{http://dx.doi.org/10.1016/0893-6080(93)90011-K}.

\bibitem[Cheng et~al.(2009)Cheng, Fu, and Wang]{cheng2009model}
Shih-Sian Cheng, Hsin-Chia Fu, and Hsin-Min Wang.
\newblock Model-based clustering by probabilistic self-organizing maps.
\newblock \emph{IEEE Transactions on Neural Networks}, 20\penalty0
  (5):\penalty0 805--826, 2009.

\bibitem[Flexer(1999)]{10.1007/978-3-540-48247-5_9}
Arthur Flexer.
\newblock On the use of self-organizing maps for clustering and visualization.
\newblock In Jan~M. {\.{Z}}ytkow and Jan Rauch, editors, \emph{Principles of
  Data Mining and Knowledge Discovery}, pages 80--88, Berlin, Heidelberg, 1999.
  Springer Berlin Heidelberg.
\newblock ISBN 978-3-540-48247-5.

\bibitem[Forest et~al.(2019)Forest, Lebbah, Azzag, and Lacaille]{Forest2019}
Florent Forest, Mustapha Lebbah, Hanane Azzag, and J{\'{e}}r{\^{o}}me Lacaille.
\newblock {Deep Embedded SOM: Joint Representation Learning and
  Self-Organization}.
\newblock In \emph{European Symposium on Artificial Neural Networks,
  Computational Intelligence and Machine Learning (ESANN 2019)}, pages 1--6,
  2019.

\bibitem[Fortuin and R{\"a}tsch(2019)]{fortuin2019deep}
Vincent Fortuin and Gunnar R{\"a}tsch.
\newblock Deep mean functions for meta-learning in gaussian processes.
\newblock \emph{arXiv preprint arXiv:1901.08098}, 2019.

\bibitem[Fortuin et~al.(2018)Fortuin, H{\"u}ser, Locatello, Strathmann, and
  R{\"a}tsch]{DBLP:journals/corr/abs-1806-02199}
Vincent Fortuin, Matthias H{\"u}ser, Francesco Locatello, Heiko Strathmann, and
  Gunnar R{\"a}tsch.
\newblock Som-vae: Interpretable discrete representation learning on time
  series.
\newblock \emph{arXiv preprint arXiv:1806.02199}, 2018.

\bibitem[Fortuin et~al.(2019)Fortuin, R{\"a}tsch, and
  Mandt]{fortuin2019multivariate}
Vincent Fortuin, Gunnar R{\"a}tsch, and Stephan Mandt.
\newblock Multivariate time series imputation with variational autoencoders.
\newblock \emph{arXiv preprint arXiv:1907.04155}, 2019.

\bibitem[Franceschi et~al.(2019)Franceschi, Dieuleveut, and
  Jaggi]{franceschi2019unsupervised}
Jean-Yves Franceschi, Aymeric Dieuleveut, and Martin Jaggi.
\newblock Unsupervised scalable representation learning for multivariate time
  series.
\newblock \emph{arXiv preprint arXiv:1901.10738}, 2019.

\bibitem[{Goodfellow} et~al.(2014){Goodfellow}, {Pouget-Abadie}, {Mirza}, {Xu},
  {Warde-Farley}, {Ozair}, {Courville}, and {Bengio}]{2014arXiv1406.2661G}
Ian~J. {Goodfellow}, Jean {Pouget-Abadie}, Mehdi {Mirza}, Bing {Xu}, David
  {Warde-Farley}, Sherjil {Ozair}, Aaron {Courville}, and Yoshua {Bengio}.
\newblock {Generative Adversarial Networks}.
\newblock \emph{arXiv e-prints}, art. arXiv:1406.2661, Jun 2014.

\bibitem[Goyal et~al.(2017)Goyal, Hu, Liang, Wang, and
  Xing]{goyal2017nonparametric}
Prasoon Goyal, Zhiting Hu, Xiaodan Liang, Chenyu Wang, and Eric~P Xing.
\newblock Nonparametric variational auto-encoders for hierarchical
  representation learning.
\newblock In \emph{Proceedings of the IEEE International Conference on Computer
  Vision}, pages 5094--5102, 2017.

\bibitem[Guo et~al.(2017)Guo, Gao, Liu, and Yin]{ijcai2017-243}
Xifeng Guo, Long Gao, Xinwang Liu, and Jianping Yin.
\newblock Improved deep embedded clustering with local structure preservation.
\newblock In \emph{Proceedings of the Twenty-Sixth International Joint
  Conference on Artificial Intelligence, {IJCAI-17}}, pages 1753--1759, 2017.
\newblock \doi{10.24963/ijcai.2017/243}.
\newblock URL \url{https://doi.org/10.24963/ijcai.2017/243}.

\bibitem[Hochreiter and Schmidhuber(1997)]{hochreiter1997long}
Sepp Hochreiter and J{\"u}rgen Schmidhuber.
\newblock Long short-term memory.
\newblock \emph{Neural computation}, 9\penalty0 (8):\penalty0 1735--1780, 1997.

\bibitem[Hudson and Cohen(2010)]{hudson2010uncertainty}
Donna~L Hudson and Maurice~E Cohen.
\newblock Uncertainty and complexity in personal health records.
\newblock In \emph{2010 Annual International Conference of the IEEE Engineering
  in Medicine and Biology}, pages 6773--6776. IEEE, 2010.

\bibitem[Jolliffe(2011)]{Jolliffe2011}
Ian Jolliffe.
\newblock \emph{Principal Component Analysis}, pages 1094--1096.
\newblock Springer Berlin Heidelberg, Berlin, Heidelberg, 2011.
\newblock ISBN 978-3-642-04898-2.
\newblock \doi{10.1007/978-3-642-04898-2_455}.
\newblock URL \url{https://doi.org/10.1007/978-3-642-04898-2_455}.

\bibitem[{Kingma} and {Welling}(2013)]{2013arXiv1312.6114K}
Diederik~P {Kingma} and Max {Welling}.
\newblock {Auto-Encoding Variational Bayes}.
\newblock \emph{arXiv e-prints}, art. arXiv:1312.6114, Dec 2013.

\bibitem[{Kohonen}(1990)]{58325}
T.~{Kohonen}.
\newblock The self-organizing map.
\newblock \emph{Proceedings of the IEEE}, 78\penalty0 (9):\penalty0 1464--1480,
  Sep. 1990.
\newblock ISSN 0018-9219.
\newblock \doi{10.1109/5.58325}.

\bibitem[Kohonen(1995)]{Kohonen1995TheAS}
Teuvo Kohonen.
\newblock The adaptive-subspace som (assom) and its use for the implementation
  of invariant feature detection.
\newblock 1995.

\bibitem[Krishnan et~al.(2016)Krishnan, Shalit, and Sontag]{krishnan2016deep}
Rahul~G Krishnan, Uri Shalit, and David Sontag.
\newblock Deep kalman filters.
\newblock 2016.

\bibitem[{Lecun} et~al.(1998){Lecun}, {Bottou}, {Bengio}, and
  {Haffner}]{726791}
Y.~{Lecun}, L.~{Bottou}, Y.~{Bengio}, and P.~{Haffner}.
\newblock Gradient-based learning applied to document recognition.
\newblock \emph{Proceedings of the IEEE}, 86\penalty0 (11):\penalty0
  2278--2324, Nov 1998.
\newblock ISSN 0018-9219.
\newblock \doi{10.1109/5.726791}.

\bibitem[Li et~al.(2018)Li, Chen, and Zhang]{DBLP:journals/corr/abs-1803-05206}
Xiaopeng Li, Zhourong Chen, and Nevin~L. Zhang.
\newblock Latent tree variational autoencoder for joint representation learning
  and multidimensional clustering.
\newblock \emph{CoRR}, abs/1803.05206, 2018.
\newblock URL \url{http://arxiv.org/abs/1803.05206}.

\bibitem[{Liu} et~al.(2015){Liu}, {Wang}, and {Gong}]{7280357}
N.~{Liu}, J.~{Wang}, and Y.~{Gong}.
\newblock Deep self-organizing map for visual classification.
\newblock In \emph{2015 International Joint Conference on Neural Networks
  (IJCNN)}, pages 1--6, July 2015.
\newblock \doi{10.1109/IJCNN.2015.7280357}.

\bibitem[L{\'o}pez-Rubio(2010)]{lopez2010probabilistic}
Ezequiel L{\'o}pez-Rubio.
\newblock Probabilistic self-organizing maps for continuous data.
\newblock \emph{IEEE Transactions on Neural Networks}, 21\penalty0
  (10):\penalty0 1543--1554, 2010.

\bibitem[MacQueen(1967)]{macqueen1967}
J.~MacQueen.
\newblock Some methods for classification and analysis of multivariate
  observations.
\newblock In \emph{Proceedings of the Fifth Berkeley Symposium on Mathematical
  Statistics and Probability, Volume 1: Statistics}, pages 281--297, Berkeley,
  Calif., 1967. University of California Press.
\newblock URL \url{https://projecteuclid.org/euclid.bsmsp/1200512992}.

\bibitem[McQueen et~al.(2004)McQueen, Hopgood, Tepper, and
  Allen]{10.1007/978-3-540-45240-9_1}
T.~A. McQueen, A.~A. Hopgood, J.~A. Tepper, and T.~J. Allen.
\newblock A recurrent self-organizing map for temporal sequence processing.
\newblock In Ahamad Lotfi and Jonathan~M. Garibaldi, editors,
  \emph{Applications and Science in Soft Computing}, pages 3--8, Berlin,
  Heidelberg, 2004. Springer Berlin Heidelberg.
\newblock ISBN 978-3-540-45240-9.

\bibitem[Min et~al.(2018)Min, Guo, Liu, Zhang, Cui, and
  Long]{IEEE:journal2018Min}
Erxue Min, Xifeng Guo, Qiang Liu, Gen Zhang, Jianjing Cui, and Jun Long.
\newblock A survey of clustering with deep learning: From the perspective of
  network architecture.
\newblock \emph{IEEE Access}, PP:\penalty0 1--1, 07 2018.
\newblock \doi{10.1109/ACCESS.2018.2855437}.

\bibitem[Moran(1950)]{10.1093/biomet/37.1-2.17}
P.~A.~P. Moran.
\newblock {NOTES ON CONTINUOUS STOCHASTIC PHENOMENA}.
\newblock \emph{Biometrika}, 37\penalty0 (1-2):\penalty0 17--23, 06 1950.
\newblock ISSN 0006-3444.
\newblock \doi{10.1093/biomet/37.1-2.17}.
\newblock URL \url{https://doi.org/10.1093/biomet/37.1-2.17}.

\bibitem[Pollard et~al.(2018)Pollard, Johnson, Raffa, Celi, Mark, and
  Badawi]{pollard2018eicu}
Tom~J Pollard, Alistair~EW Johnson, Jesse~D Raffa, Leo~A Celi, Roger~G Mark,
  and Omar Badawi.
\newblock The eicu collaborative research database, a freely available
  multi-center database for critical care research.
\newblock \emph{Scientific data}, 5, 2018.

\bibitem[Razavi et~al.(2019)Razavi, Oord, and Vinyals]{razavi2019generating}
Ali Razavi, Aaron van~den Oord, and Oriol Vinyals.
\newblock Generating diverse high-fidelity images with vq-vae-2.
\newblock \emph{arXiv preprint arXiv:1906.00446}, 2019.

\bibitem[{Tirunagari} et~al.(2014){Tirunagari}, {Poh}, {Aliabadi}, {Windridge},
  and {Cooke}]{7008682}
S.~{Tirunagari}, N.~{Poh}, K.~{Aliabadi}, D.~{Windridge}, and D.~{Cooke}.
\newblock Patient level analytics using self-organising maps: A case study on
  type-1 diabetes self-care survey responses.
\newblock In \emph{2014 IEEE Symposium on Computational Intelligence and Data
  Mining (CIDM)}, pages 304--309, Dec 2014.
\newblock \doi{10.1109/CIDM.2014.7008682}.

\bibitem[Tokunaga and Furukawa(2009)]{TOKUNAGA200982}
Kazuhiro Tokunaga and Tetsuo Furukawa.
\newblock Modular network som.
\newblock \emph{Neural Networks}, 22\penalty0 (1):\penalty0 82 -- 90, 2009.
\newblock ISSN 0893-6080.
\newblock \doi{https://doi.org/10.1016/j.neunet.2008.10.006}.
\newblock URL
  \url{http://www.sciencedirect.com/science/article/pii/S0893608008002335}.

\bibitem[van~den Oord et~al.(2017)van~den Oord, Vinyals, and
  Kavukcuoglu]{DBLP:journals/corr/abs-1711-00937}
A{\"{a}}ron van~den Oord, Oriol Vinyals, and Koray Kavukcuoglu.
\newblock Neural discrete representation learning.
\newblock \emph{CoRR}, abs/1711.00937, 2017.
\newblock URL \url{http://arxiv.org/abs/1711.00937}.

\bibitem[Vikram et~al.(2018)Vikram, Hoffman, and Johnson]{vikram2018loracs}
Sharad Vikram, Matthew~D Hoffman, and Matthew~J Johnson.
\newblock The loracs prior for vaes: Letting the trees speak for the data.
\newblock \emph{arXiv preprint arXiv:1810.06891}, 2018.

\bibitem[Voegtlin(2002)]{VOEGTLIN2002979}
Thomas Voegtlin.
\newblock Recursive self-organizing maps.
\newblock \emph{Neural Networks}, 15\penalty0 (8):\penalty0 979 -- 991, 2002.
\newblock ISSN 0893-6080.
\newblock \doi{https://doi.org/10.1016/S0893-6080(02)00072-2}.
\newblock URL
  \url{http://www.sciencedirect.com/science/article/pii/S0893608002000722}.

\bibitem[Xiao et~al.(2017)Xiao, Rasul, and
  Vollgraf]{DBLP:journals/corr/abs-1708-07747}
Han Xiao, Kashif Rasul, and Roland Vollgraf.
\newblock Fashion-mnist: a novel image dataset for benchmarking machine
  learning algorithms.
\newblock \emph{CoRR}, abs/1708.07747, 2017.
\newblock URL \url{http://arxiv.org/abs/1708.07747}.

\bibitem[Xie et~al.(2015)Xie, Girshick, and
  Farhadi]{DBLP:journals/corr/XieGF15}
Junyuan Xie, Ross~B. Girshick, and Ali Farhadi.
\newblock Unsupervised deep embedding for clustering analysis.
\newblock \emph{CoRR}, abs/1511.06335, 2015.
\newblock URL \url{http://arxiv.org/abs/1511.06335}.

\bibitem[Yang et~al.(2016)Yang, Fu, Sidiropoulos, and
  Hong]{DBLP:journals/corr/YangFSH16}
Bo~Yang, Xiao Fu, Nicholas~D. Sidiropoulos, and Mingyi Hong.
\newblock Towards k-means-friendly spaces: Simultaneous deep learning and
  clustering.
\newblock \emph{CoRR}, abs/1610.04794, 2016.
\newblock URL \url{http://arxiv.org/abs/1610.04794}.

\end{thebibliography}

\newpage

\appendix

\section*{Appendix}

\section{Self-Organizing Maps}
\label{sec:self_organizing_maps}
Among unsupervised learning algorithms supporting a neighborhood structure between clusters, 
Kohonen's self-organizing map (SOM) \citep{58325} is one of the most
popular models. It is comprised of $K$ neurons connected to form a discrete topological structure. The data points are projected 
onto this topographic map which locally approximates the data manifold. Usually the map is chosen as a finite two-dimensional region where neurons 
are arranged in a regular hexagonal or rectangular grid. In our work we use a rectangular grid, $M \subseteq \mathbb{N}^2$, because of its simplicity and its ease of visualization. 
Each neuron $m_{ij}$ at position $(i,j)$ of the grid, for $i,j = 1, \dots, \sqrt{K}$, corresponds to a centroid vector $\mu_{i,j}$ in the input space. 
The centroids are tied by a neighborhood relation, here defined as $N\left(\mu_{i, j}\right)=\left\{\mu_{i-1, j}, \mu_{i+1, j}, \mu_{i, j-1}, \mu_{i, j+1}\right\}$.

Given a random initialization of the centroids, the SOM algorithm randomly selects an input $x_i$ and updates both its closest centroid $\mu_{i,j}$ and its neighbors $N\left(\mu_{i, j}\right)$ to move them closer to $x_i$. The algorithm (\ref{algSOM}) then iterates these steps until convergence.

\begin{algorithm}
\caption{Self-Organizing Maps}
\label{algSOM}
\begin{algorithmic} 
\REQUIRE $0<\alpha(t)<1; \;\;\;\; \lim _{t \rightarrow \infty} \sum \alpha(t) \rightarrow \infty; \;\;\;\; \lim _{t \rightarrow \infty} \sum \alpha^{2}(t)<\infty; $
\REPEAT
\STATE At each iteration $t$, present an input $x(t)$ and select the winner, $$\nu(t)=\arg \min _{k \in \Omega}\left\|\mathbf{x}(t)-\mathbf{w}_{k}(t)\right\|$$
\STATE Update the weights of the winner and its neighbors, $$\Delta \mathbf{w}_{k}(t)=\alpha(t) \eta(\nu, k, t)\left[\mathbf{x}(t)-\mathbf{w}_{\nu}(t)\right]$$
\UNTIL{the map converges}
\end{algorithmic}
\end{algorithm}

The range of SOM applications includes high dimensional data visualization, clustering, image and video processing, density or spectrum profile 
modeling, text/document mining, management systems and gene expression data analysis. %

\newpage

\section{Data and implementation details}
\label{sec:details}

\subsection{Datasets}
\label{subsec:datasets}
\begin{itemize}
    \item \textbf{MNIST:} It consists of 70,000 handwritten digits of 28-by-28 pixel size. Digits range from 0 to 9, yielding 10 patterns in total. The digits have been size-normalized and centered in a fixed-size image \cite{726791}. 
    \item \textbf{Fashion MNIST:} A dataset of Zalando's article images consisting of a training set of 60,000 examples and a test set of 10,000 examples \cite{DBLP:journals/corr/abs-1708-07747}. Each example is a 28$\times$28 grayscale image, associated with a label from 10 classes. 
    \item \textbf{eICU:} We use vital sign/lab measurements of intensive care unit (ICU) patients resampled to a 1-hour
                         based grid using forward filling and filling with population statistics from the training set if no measurements were available
                         prior to the time point. From all available ICU stays, we excluded stays which were shorter than 3 days, longer than 30 days, or
                         which had at least one gap in the continuous vital sign monitoring, which we define as an interval between two heart rate measurements of 
                         at least 1 hour. This yielded $N=10 559$ ICU stays from the eICU database \cite{pollard2018eicu}. We included $d_\text{vitals}=14$ vital sign 
                         variables and $d_\text{lab}=84$ lab measurement variables, giving an overall
                         data dimension of $d=98$. The last 72 hours of these multivariate time series were used for the experiments. As labels, we use a variant of 
                         the current dynamic APACHE physiology score (APACHE-0) as well as the worst APACHE score in the next $\{6,12,24\}$ hours (APACHE-6/12/24), and
                         mortality risk in the next 24 hours. Only those variables from the APACHE score definition which are recorded in the eICU 
                         database were taken into account for its definition.
                         
\end{itemize}
Each dataset is divided into training, validation and test sets for both our models and the baselines.

\subsection{Implementation}
\label{subsec:implementation}
In implementing our models we focused on retaining a fair comparison with the baselines. Hence, we decided to use a standard network 
structure, with fully connected layers of dimensions $d-500-500-2000-l$, to implement both the VAEs and the AEs. On the static data, the latent dimension $l$ is set to $100$ for the VAE, and to $10$ for the AEs.
Since the prior in the VAE restricts the latent embeddings to be compact, it also requires more dimensions to learn a 
meaningful latent space. We observed that, providing the AE models with a higher-dimensional latent space, as used for the VAE, resulted in a dramatic 
decrease of performance (see Appendix \ref{subsec:latent_space_ae}).
We set the number of clusters to $64$ for the image data and to $256$ for the ICU time series for both the baselines and our models. 
We choose our SOM grid to be the 2-dimensional surface of a 3-dimensional torus in all experiments.%
The neighborhood is chosen to contain one neighboring cluster in each direction (up, down, left, right).
For other grid dimensions we refer to the Appendix \ref{subsec:num_clusters}. 

\newpage

\section{Hyperparameter tuning and model robustness}
\label{subsec:HP}
\begin{figure}[h!]
\begin{center}
\subfloat{\includegraphics[width=0.49\textwidth]{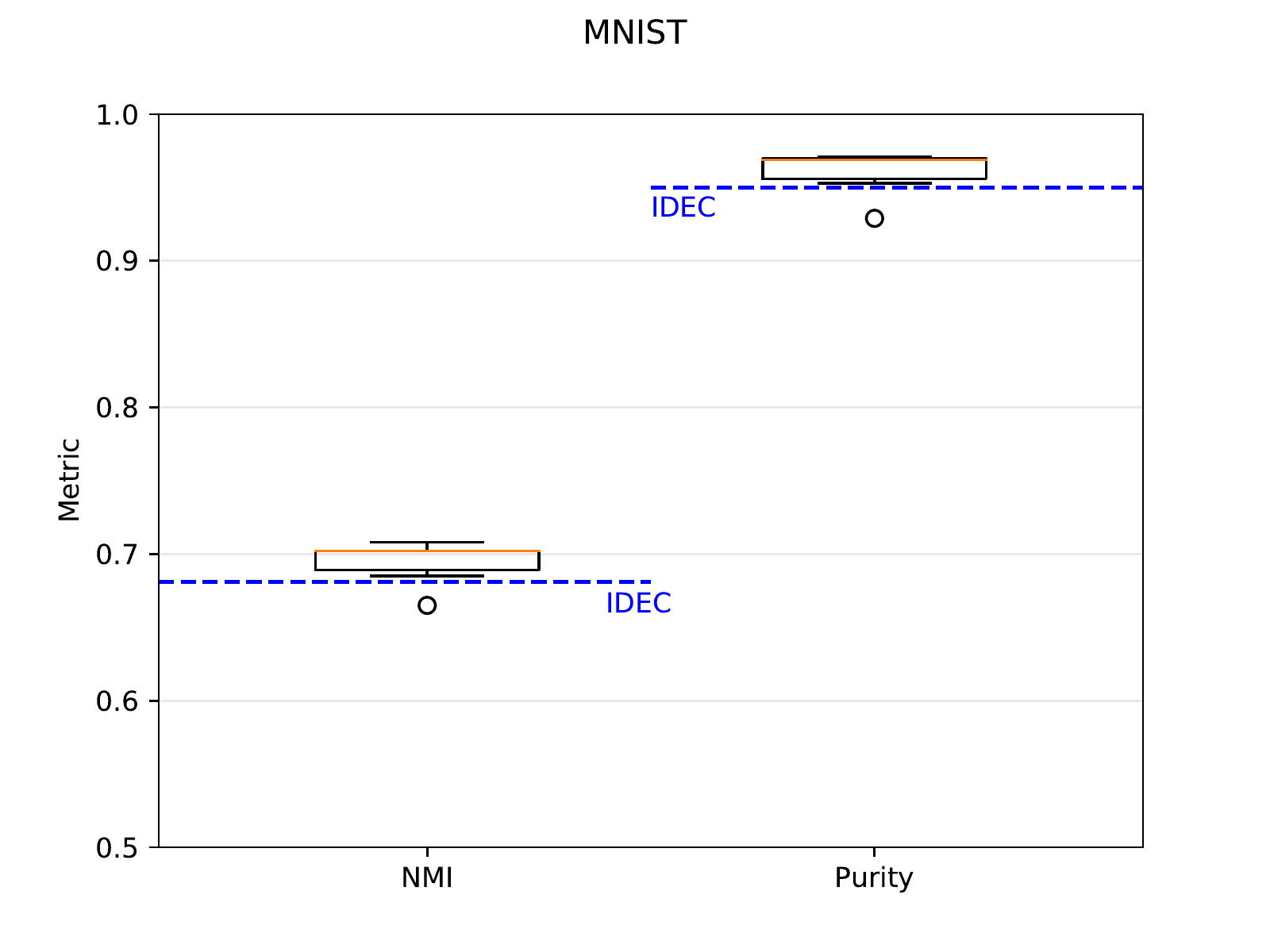}}
\hfill
\subfloat{\includegraphics[width=0.49\textwidth]{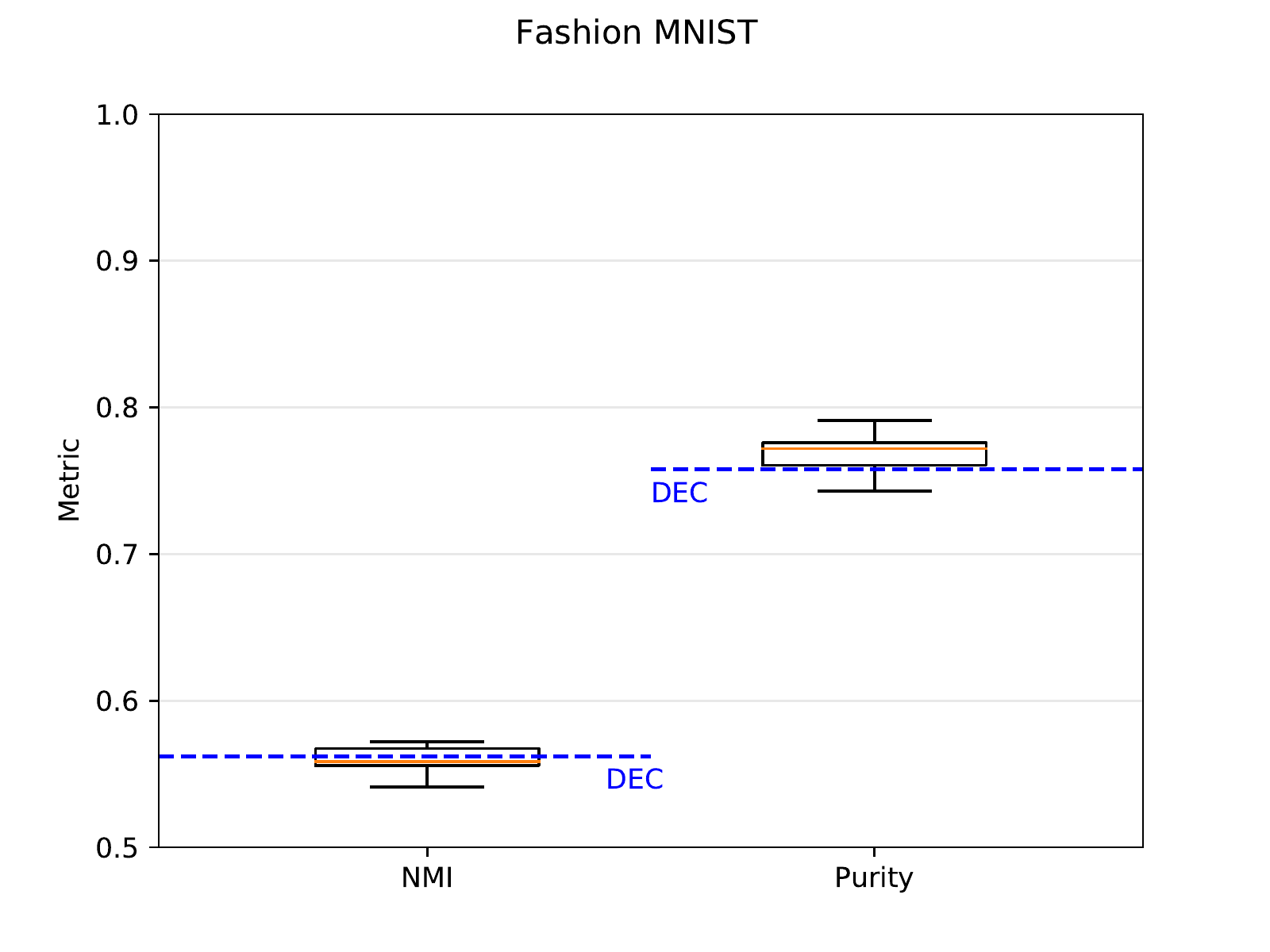}}
\end{center}
\caption{Robustness of DPSOM performance with respect to different choices of hyperparameters $\beta$ and $\gamma$ on 
        the image clustering task. 10 replicates were drawn by choosing $\gamma$ randomly in $[10,30]$, $\beta \in [0,1]$ and using a different random 
        seed for each experiment. As a reference we plot as a dotted blue line the result of the best performing competitor, which is 
        IDEC and DEC for the MNIST/Fashion-MNIST data-sets, respectively.}
     \label{HP}
\end{figure}

Our models DPSOM/T-DPSOM contain two hyperparameters, $\beta$ and $ \gamma$, which have to be fine-tuned by the user to their data set.
In this section, we define heuristics on how to choose them and we assess the robustness of the DPSOM model to different hyperparameter settings
for the image data-sets. In practice we have found that $\gamma$ should be chosen such that the magnitude of the VAE reconstruction loss is 
at least 10x than $l_\text{CAH}$ throughout training. $\beta$ should be ideally chosen such that the S-SOM loss has a similar magnitude as the CAH loss.
However, as long as this ratio is approximately maintained, it can be adjusted by the user to their desired level of spatial
coherence, and we found its exact value is irrelevant to the raw clustering performance. These simple heuristics for choosing the hyperparameters are 
completely unsupervised as they depend only on the magnitude of different loss components. They do not rely on ground-truth label information or any expensive 
optimization, and we found that they worked well in practice for all datasets. 
Table \ref{table_HP} summarizes the hyper-parameter settings as well as other training parameters for both static and temporal data.
Figure \ref{HP} shows the NMI and purity results achieved by the DPSOM model for random values of $\beta \in [0,1]$ and $ \gamma \in [10,30]$ for both MNIST and fMNIST. 
We observe that the performance of DPSOM is robust to different choices of the hyperparameters, and it generally outperforms its main competitor, 
DEC/IDEC, which was tuned in a supervised way, for all metrics but NMI on Fashion-MNIST.

\begin{table}[h!]
  \caption{Hyperparameter settings and training parameters for the static data sets (MNIST and fMNIST) and the temporal data set (eICU)}
    \label{table_HP}
  \begin{center}
    \resizebox{\linewidth}{!}{
      \begin{tabular}{cccccc}
        \textbf{Task} & \textbf{Image clustering (MNIST/fMNIST)} & \textbf{Temporal clustering/forecasting (eICU)} \\
        \midrule
      $\gamma$ (ELBO)   & $ 20$  & $ 50 $ \\
      $\beta$ (Soft-SOM)   & $0.25 $  / $0.4 $& $10$\\
      $\alpha$ (CAH parameter) & $10$&$10$\\
      SOM dimension & $8 \times 8$ & $16 \times 16$ \\
      Batch size & $300$ & $300$ \\
      Num. of epochs   & $300$ & $100$ \\
      Latent dimension & $100$& $50$ / $100$\\
      Learning rate & $0.001$ &$0.001$\\
      Dropout &$0.4 $  &$0.5 $ \\
      \end{tabular}
    }
  \end{center}
\end{table}

\FloatBarrier

\section{Effect of other parameters and training details}

\subsection{Performance for different numbers of clusters}
\label{subsec:num_clusters}
\begin{figure}[h!]
  \begin{center}
  \subfloat{\includegraphics[width=0.45\textwidth]{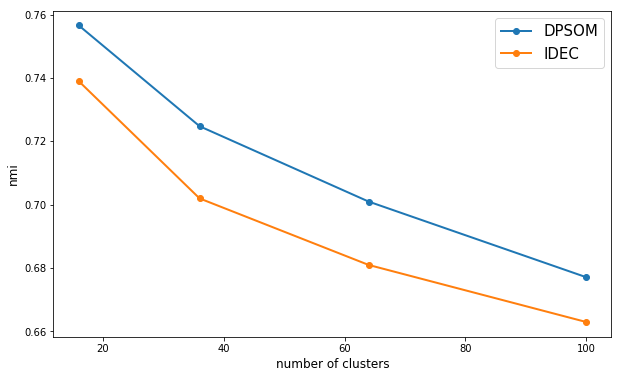}}
    \hfill
  \subfloat{\includegraphics[width=0.45\textwidth]{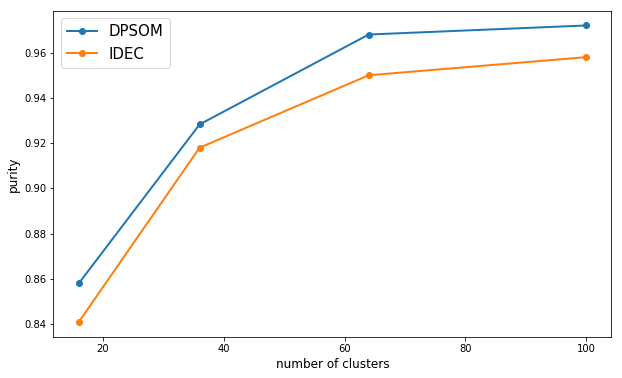}}
  \end{center}
  \caption{NMI (left) and purity (right) clustering performance of DPSOM and IDEC when varying the number of clusters on the MNIST test set.}
  \label{fig:clusters}
\end{figure}

We evaluate the NMI and purity clustering performance of our model (DPSOM) while varying the number of clusters on the MNIST test set. 
Since IDEC represents its main competitor we only include this model in this analysis. Figure~\ref{fig:clusters} shows that DPSOM outperforms IDEC
for all chosen number of clusters. It is particularly interesting to observe that NMI decreases with an increasing number of clusters in both models. 
We suspect this is because the entropy of the clustering increases as the number of clusters is increased.

\FloatBarrier
\FloatBarrier

\subsection{Effect of latent space dimension in DEC}
\label{subsec:latent_space_ae}
To provide a fair comparison with the DEC baseline, we evaluated the DEC model for different latent space dimensions. Table
\ref{table:l} shows that the AE, used in the DEC model, performs better when a lower-dimensional latent space is used. 

\begin{table}[ht!]
\caption{Mean/Standard error of NMI and purity of the DEC model on the MNIST test set, across 10 runs with different random model initializations. We 
         use 64 clusters and different latent space dimensions.}
\label{table:l}
\begin{center}
\begin{tabular}{@{}lccc@{}}\toprule
Latent dimension & Purity  & NMI
\\ \midrule
 $l= 10$& $ 0.950 \pm 0.001$ & $0.681 \pm 0.001$\\
  $l= 100$ & $ 0.750 \pm 0.001$  &  $0.573 \pm 0.001 $\\
 \bottomrule
\end{tabular}
\end{center}
\end{table}
\FloatBarrier

\subsection{Performance improvement over training}
\begin{figure}[h!]
	\begin{center}
		\subfloat{\includegraphics[width=0.49\linewidth]{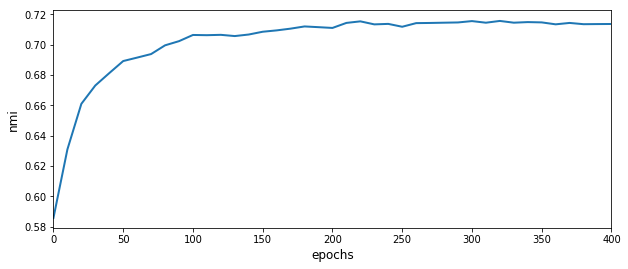}}
		\hfill
		\subfloat{\includegraphics[width=0.49\linewidth]{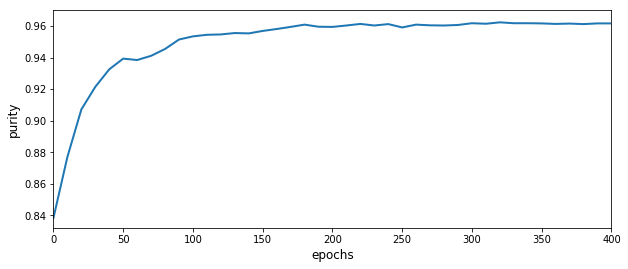}}
	\end{center}
	\caption{NMI (top) and Purity (bottom) performance of DPSOM over the number of epochs, on the MNIST test set.}
	\label{fig:temp}
\end{figure}

After obtaining the initial configuration of the SOM structure, both clustering and feature extraction using the VAE
are trained jointly. To illustrate that our architecture improves clustering performance over the initial configuration, we plotted NMI and Purity 
against the number of training epochs in Figure~\ref{fig:temp}. We observe that the performance converges when increasing the number of
epochs.

\newpage

\section{Computational considerations and training time}
\label{sec:comp-issues}

For each epoch, we require a scan over all pairs of centroids and embedded data points to compute the $\{ t_{ij} \}$, which 
has a complexity proportional to  $\mathcal{O}(k \cdot n_\text{batch})$, where $k$ is the number of centroids and $n_\text{batch}$ is the batch size.
The average over the data set can either be computed exactly in a streaming fashion once every epoch or approximated using a running estimate. 

In practice, the model fitting time of T-DPSOM on the eICU training set, which contains 517,000 time points of dimension $d=98$, using a 
SOM grid of 16x16, to convergence is 28 minutes, using 1 GPU. On the full data-set, SOM pre-training, AE pre-training, 
main training phase, and prediction fine-tuning contributed 2, 8, 75 , 15 \% respectively to the overall training
time. Table \ref{table:trainingtime} shows how the training time scales with the training set size, and 
table \ref{table:trainingtime-nclusters} shows how training time scales with the number of clusters. We observe 
empirically that if $n$ is increased by a factor of 20, training time only increases by a factor of 8.7. The 
dependence on $K$ is even weaker, as we increase $K$ by a factor of 25 from 16 to 400, training
time only increases by 1.7x, suggesting that T-DPSOM could be used with an even finer SOM grid resolution than shown
in the main paper. All experiments were performed using a Geforce GTX 1080 GPU together with a Xeon E5-2630 CPU on Tensorflow 2.0.

\begin{table}[ht!]
\caption{Training time (per epoch) of different phases of the T-DPSOM training algorithm on 
         the eICU data set, as the size of the training data is varied. A SOM dimension of 16x16 with 
         256 clusters was used.}
\label{table:trainingtime}
\begin{center}
\resizebox{\linewidth}{!}{
\begin{tabular}{@{}lcccccc@{}}\toprule
\textbf{Training data size} & 5\% (n=359) & 10\% (n=718) & 20\% (n=1436) & 50\% (n=3590) & 100\% (n=7180)
\\ \midrule
SOM init time/epoch [s] & 0.24 & 0.29 & 0.38 & 0.71 & 1.27 \\
AE pretrain time/epoch [s] & 0.50 & 0.68 & 1.04 & 2.30 & 4.47 \\
Train time/epoch [s] & 1.24 & 1.78 & 2.83 & 6.38 & 12.46 \\ 
Pred finetune time/epoch [s] & 0.23 & 0.27 & 0.36 & 0.67 & 1.22 \\
Total training time [s] & 191.5 & 261.5 & 397.2 & 863.2 & 1662.2 
\\ \bottomrule
\end{tabular}
}
\end{center}
\end{table}

\begin{table}[ht!]
\caption{Training time (per epoch) of different phases of the T-DPSOM training algorithm on
         the eICU data set, as the SOM dimension is varied. The entire training set (n=7180) was used.}
\label{table:trainingtime-nclusters}
\begin{center}
\resizebox{\linewidth}{!}{
\begin{tabular}{@{}lcccccc@{}}\toprule
\textbf{SOM dim.} & 4x4 (K=16) & 8x8 (K=64) & 12x12 (K=144) & 16x16 (K=256) & 20x20 (K=400)
\\ \midrule
SOM init time/epoch [s] & 1.03 & 1.08 & 1.14 & 1.27 & 1.84 \\
AE pretrain time/epoch [s] & 4.36 & 4.40 & 4.42 & 4.47 & 4.94 \\
Train time/epoch [s] & 8.42 & 9.20 & 10.58 & 12.46 & 14.86 \\ 
Pred finetune time/epoch [s] & 1.11 & 1.15 & 1.17 & 1.22 & 1.71 \\
Total training time [s] & 1225.3 & 1313.8 & 1458.3 & 1662.2 & 2031.9 
\\ \bottomrule
\end{tabular}
}
\end{center}
\end{table}

\newpage

\section{Further experiments on ICU data}

\subsection{Randomly sampled patient trajectories on the SOM grid}
\label{subsec:icu_uncertainty}

\begin{figure}[h!]
\centering
  \includegraphics[width=1.0\textwidth]{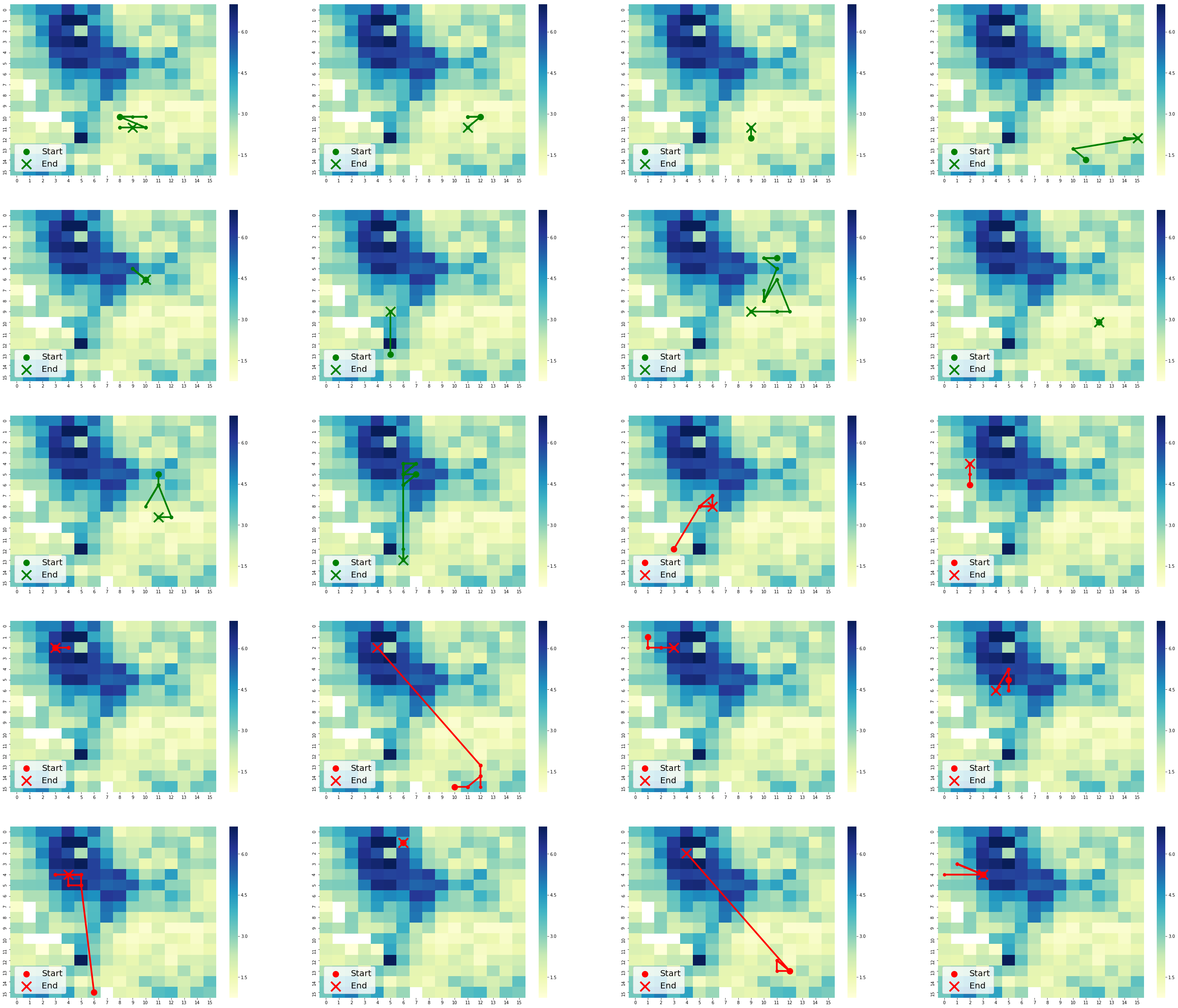}
  \caption{Randomly sampled T-DPSOM trajectories, from patients dying at the end of the ICU stay, as well as surviving
           patients. Superimposed is a heatmap which displays the mean APACHE score of all time points assigned to each cluster. We observe that
           trajectories of dying patients are often in different locations of the map compared to surviving patients, in particular in those
           regions enriched for high APACHE scores, which conforms with clinical intuition.}
  \label{fig:traj}
\end{figure}

T-DPSOM induces trajectories on the 2D SOM grid which can be easily visualized. 
Fig.~\ref{fig:traj} shows $20$ randomly sampled patient trajectories obtained by our model. 
Trajectories ending in the death of the patient are shown in red, while patients dispatched alive from the ICU are shown in green.
A different model initialization was chosen compared to the trajectories in the main paper, so the heatmap
patterns are different.

\newpage

\subsection{Cluster enrichment for current APACHE score and mortality risk}
\label{subsec:state_representation_icu}

\begin{figure}[ht!] 
  \begin{center}
  \subfloat[Current APACHE score]{\includegraphics[width=0.4\textwidth]{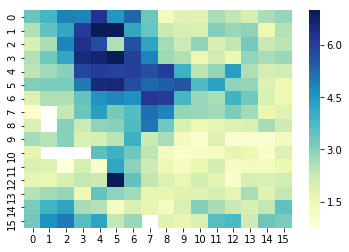}\label{fig_score}}
  \hfill
  \subfloat[Mortality risk in the next 24 hours]{\includegraphics[width=0.4\textwidth]{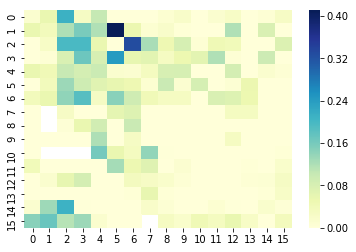}\label{fig_mort}}
   \end{center}
  \caption{Heatmaps of enrichment in mortality risk in the next 24 hours as well as the current dynamic APACHE score, superimposed on the discrete
           2D grid learned by the T-DPSOM model.}
\label{fig_states}
\end{figure}

In Fig.~\ref{fig_states}, heatmaps (colored according to enrichment in the current APACHE score, as well as the mortality risk in the next 24 hours) 
show compact enrichment structures. The T-DPSOM model succeeds in discovering a meaningful and smooth neighborhood structure with respect
to APACHE score enrichment. For mortality, it distinguishes risk profiles with practically zero mortality risk from very high mortality risk
in the next 24 hours (reaching up to 30 - 40~\%) in different regions of the map, even though it is trained in a purely unsupervised fashion, 
which is a remarkable result. Also, regions with high mortality risk often coincide with those enriched in high APACHE scores, 
which conforms with clinical intuition. A different model initialization was chosen compared to the trajectories in the main paper, 
so the heatmap patterns are different.

\end{document}